\documentclass[lettersize,journal]{IEEEtran}
\usepackage{amsmath,amsfonts}
\usepackage{algorithmic}\usepackage[T1]{fontenc}
\usepackage[ruled,vlined]{algorithm2e}
\usepackage{ragged2e}
\usepackage{cite}
\usepackage{tabularx}
\usepackage{multicol}
\usepackage{multirow}
\usepackage{amssymb}
\usepackage{tikz}
\usepackage{nicematrix}

\usepackage{pifont}
\usepackage[normalem]{ulem}
\usepackage{bm}
\usepackage{array}
\usepackage[caption=false,font=normalsize,labelfont=sf,textfont=sf]{subfig}
\usepackage{textcomp}
\usepackage{amsmath}
\usepackage{siunitx} 
\usepackage{stfloats}
\usepackage{booktabs}
\usepackage{makecell}

\usepackage{url}
\usepackage{verbatim}
\usepackage{graphicx}
\def\BibTeX{{\rm B\kern-.05em{\sc i\kern-.025em b}\kern-.08em
    T\kern-.1667em\lower.7ex\hbox{E}\kern-.125emX}}
\usepackage{balance}
\begin{document}

\title{Feature Recalibration Based Olfactory–Visual Multimodal Model for Enhanced Rice Deterioration Detection}
\author{~Rongqiang~Zhao, ~Hengrui Hu, ~Yijing Wang, ~Mingchun Sun$^{\ast}$\thanks{*Corresponding author: sunmingchun@stu.hit.edu.cn.}, Jie Liu, \textit{Fellow, IEEE}
\thanks{The authors are with the Faculty of Computing, Harbin Institute of Technology, Harbin 150001, China.}
\thanks{The authors are with the National Key Laboratory of Smart Farm Technologies and Systems, Harbin 150001, China.}}

\markboth{}%
{Feature Recalibration Based Olfactory–Visual Multimodal Model for Enhanced Rice Deterioration Detection}

\maketitle

\begin{abstract}
Multimodal methods are widely used in rice deterioration detection, which exhibit limited capability in representing and extracting fine-grained abnormal features. Moreover, these methods rely on devices, such as hyperspectral cameras and mass spectrometers, increasing detection costs and prolonging data acquisition time. To address these issues, we propose a feature recalibration based olfactory–visual multimodal model for enhanced rice deterioration detection. The fine-grained deterioration embedding constructor (FDEC) is proposed to reconstruct the labeled multimodal embedded-feature dataset, enhancing sample representation. The fine-grained deterioration recalibration attention network (FDRA-Net) is proposed to emphasize signal variations and increase sensitivity to fine-grained deterioration on the rice surface. Compared with SS-Net, the proposed method improves the classification accuracy by 8.67\%, with an average improvement of 11.51\% over other traditional baseline models, while simultaneously simplifying the detection procedure. Furthermore, field detection demonstrates the advantages of accuracy and operational simplicity. The proposed method can also be extended to other agrifood in agriculture and food industry.

\end{abstract}



\section{Introduction}
\IEEEPARstart{R}{ice} grains are susceptible to biological and chemical factors that induce fine-grained deterioration cues, making rapid and reliable rice deterioration detection critically important during rice storage\cite{[1]TIM}. Conventional sensory evaluation methods rely heavily on human experience and subjective judgment, and the results are easily influenced by interobserver variability, fatigue, and environmental conditions. Common physicochemical inspection techniques typically depend on high-precision instruments such as gas chromatographs and infrared spectrometers\cite{[2]}. Although these methods provide high analytical accuracy, they often involve expensive equipment, complex operational procedures, and long inspection cycles\cite{[3]}. Consequently, neither approach is well suited for large-scale rice storage scenarios, highlighting the need to reduce detection costs and simplify procedures while maintaining reliable detection accuracy\cite{[4]TIM}.

Artificial intelligence methods based on olfactory sensing have received growing attention in agrifood detection within the agriculture and food industry, offering new technical pathways for rice deterioration detection in large-scale storage scenarios\cite{[5]TII}. These methods are widely adopted due to their nondestructive nature and high sensitivity to volatile metabolites. The gas-sensitive materials in the electronic nose (e-nose) exhibit characteristic variations in response to changes in gas type and concentration, thereby generating discriminative signals\cite{[6]TIM}. In conjunction with e-nose, a wide range of machine learning and deep learning methods are applied in detection tasks, including multilayer perceptron (MLP), spiking neural network (SNN), convolutional neural network (CNN)\cite{[7]TIM}, long short-term memory network (LSTM)\cite{[8]TIM}, and Transformer-based model\cite{[9]}. However, olfactory information alone primarily reflects odor variations and fails to characterize surface degradation and other visual attributes. Moreover, under complex storage conditions, such methods are susceptible to sensor drift, which introduces limitations in classification robustness\cite{[10]TIM, [11]TIM}.

Multimodal detection that fuses olfactory sensor data with visual information has attracted increasing attention from researchers. The olfactory modality reflects variations in volatile components, whereas the visual modality provides a direct and intuitive representation of surface texture changes, making the two sources of information complementary at the representation level. Men et al. propose an interleaved attention convolutional compression network (IACCN) that integrates e-nose data with hyperspectral data, achieving inter-modal information interaction and redundancy compression through an interleaved attention mechanism, thereby improving feature utilization efficiency and classification performance\cite{[12]IACNN}. Shi et al. further propose a collaborative fusion strategy that feeds multisource data acquired from the e-nose and hyperspectral imaging into deep networks for feature extraction, combined with improved machine learning classifiers to enhance inter-class separability and recognition stability\cite{[13]GE-ELM}. More recently, Shi et al. design a gas–spectral bimodal fusion framework based on electronic nose and hyperspectral systems, in which a self-selective convolutional structure adaptively focuses on key features, enabling effective discrimination of rice quality across different storage periods. By fusing olfactory and visual features, multimodal methods generally achieve more stable and more discriminative deterioration detection results than single-modality methods\cite{[14], [15]}.

Despite the demonstrated advantages of multimodal fusion in rice deterioration detection, existing methods still exhibit certain limitations. First, their capability to represent and extract fine-grained abnormal features remains limited, which makes it difficult to reliably capture subtle and sparse deterioration cues. Second, these methods rely on specialized devices, such as hyperspectral cameras and mass spectrometers, during data acquisition. This dependence not only increases detection costs but also prolongs data acquisition and imaging time, thereby reducing the practicality of large-scale deployment. Therefore, improving fine-grained feature modeling while reducing the reliance on costly and time-consuming devices remains a critical challenge for multimodal rice deterioration detection.

To address these issues, a multimodal enhanced rice deterioration detection method is proposed. The method makes progress in improving detection accuracy and simplifying the detection procedure. The fine-grained deterioration embedding constructor (FDEC) is proposed to reconstruct the labeled multimodal embedded-feature dataset, enhancing sample representation. The fine-grained deterioration recalibration attention network (FDRA-Net) is proposed to emphasize signal variations and increase sensitivity to fine-grained deterioration on the rice surface. The contributions of this study are summarized as follows:
\begin{enumerate}
\item{
We establish the multimodal enhanced rice deterioration detection framework to ensure the accuracy, separability, and cost-effectiveness of data acquisition.
}
\item{
We propose the FDEC to reconstruct a labeled multimodal embedded-feature dataset, thereby enhancing sample representation.
}
\item{
We propose the FDRA-Net to emphasize signal variations and increase sensitivity to fine-grained deterioration on the rice surface.
}
\end{enumerate}

The organization of this article is as follows. In Section II, the methodology and theory are elaborated in detail. In Section III, the experimental results are presented and discussed, and conclusions are drawn in Section IV.


\section{Methodology}

\begin{figure*}[t]
\centering
\includegraphics[width=7.0in]{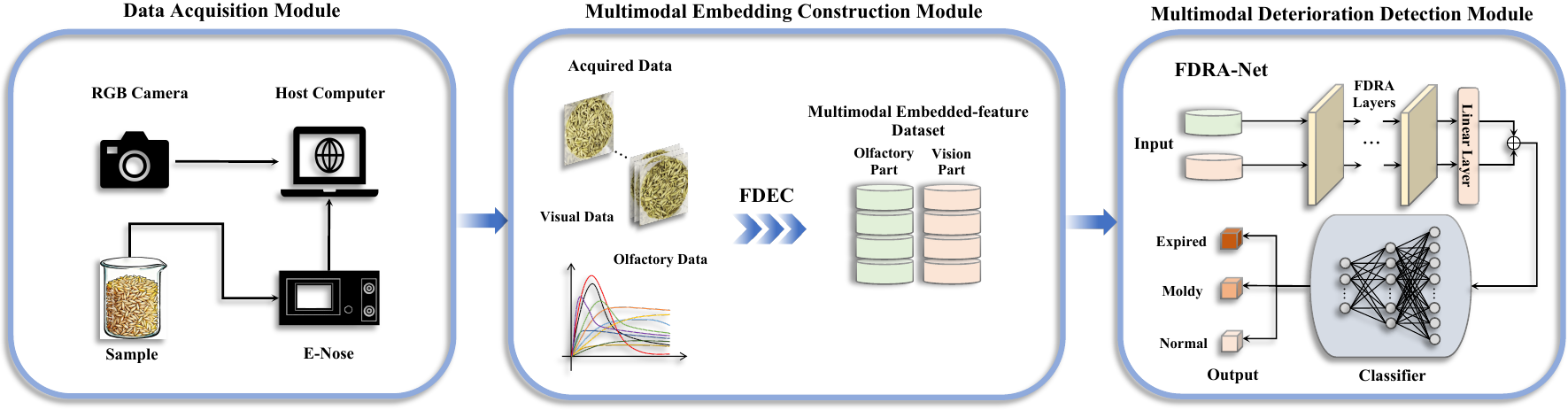}
\caption{Multimodal Enhanced Rice Deterioration Detection Framework.}
\label{proposed_framework}
\end{figure*}

The multimodal enhanced rice deterioration detection framework consists of three functional modules: the data acquisition module, the multimodal embedding construction module, and the multimodal deterioration detection module. In the data acquisition module, the host computer receives data collected by the RGB camera and the e-nose. The acquired data are subsequently processed by the FDEC within the multimodal embedding construction module to reconstruct the labeled multimodal embedded-feature dataset. The FDRA-Net in the multimodal deterioration detection module then accesses the dataset to perform rice deterioration detection. An overview of the proposed framework is presented in Fig. \ref{proposed_framework}.

\subsection{Data  Acquisition Module}
The data acquisition module primarily consists of an RGB camera and an e-nose. The RGB camera (SARGO G20, SARGO E-Commerce Co., Ltd., Shenzhen, China) is used to collect visual data of rice grains from a top-down perspective. The e-nose (PEN3, AIRSENSE Analytics, Schwerin, Mecklenburg, Germany) is applied to acquire olfactory data of the rice grains. The RGB camera and the e-nose are connected to the host computer via USB interfaces, and the host computer ensures real-time processing of the acquired data.

The data acquisition environment maintains a temperature of approximately 24 $^\circ\mathrm{C}$ and a relative humidity of about 40\%\cite{[16], [17]}. The mass of each rice sample is kept consistent and strictly controlled at 60 g, and the sample is placed in a beaker. The RGB camera is configured at 4K and 24 FPS and records the rotating rice sample from a top-down perspective, producing three video segments of 25 s each. To ensure diversity in the visual data, the sample is thoroughly stirred during the intervals between video recordings to alter its surface visual texture. The e-nose is then used to collect the olfactory data, with a recording duration of 600 s.

\begin{figure}[h]
\centering
\includegraphics[width=3.5in]{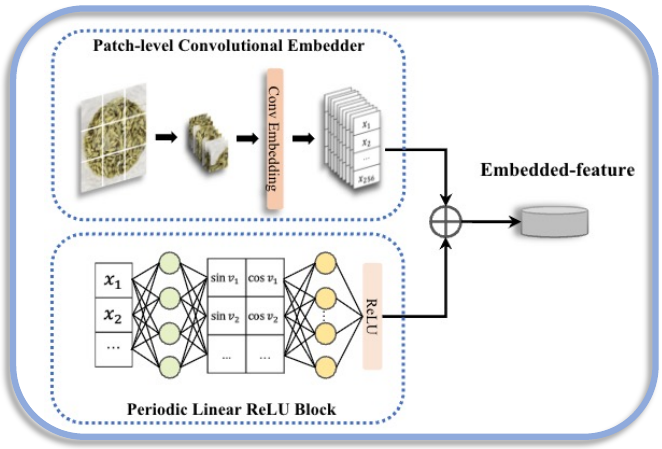}
\caption{The Structure of FDEC.}
\label{FDEC_structure}
\end{figure}

\subsection{Multimodal Embedding Construction Module}
The raw RGB images and e-nose signals exhibit substantial heterogeneity in both form and statistical characteristics. The RGB images contain sparse and spatially irregular mold cues, whereas the olfactory signals manifest as temporal responses with varying noise levels and unstable rising phases. Directly feeding such heterogeneous signals into a multimodal network often leads to inconsistent feature representations and redundant information. To mitigate these issues, we propose the FDEC, which standardizes and reorganizes the two modalities prior to their respective branch processing. Instead of enforcing dimensional alignment between the visual and olfactory data, FDEC independently embeds each modality and arranges the resulting representations within a unified multimodal dataset structure. This design effectively reduces modality-specific noise, enhances discriminative local variations, and provides a more stable foundation for subsequent multimodal feature extraction.

FDEC constructs a structured multimodal preprocessing and embedding pipeline to process visual and olfactory data and convert them into a uniformly managed embedding structure as shown in Fig. \ref{FDEC_structure}. For the olfactory modality, FDEC first extracts the stable response phase from the raw e-nose signals and then organizes them into an input $\mathbf{S}$ suitable for the periodic linear ReLU (PLR)\cite{[18]PLR} module by applying feature emphasizing and normalization. The PLR module, acting as a numerical embedding function $f_{\text{PLR}}(\cdot)$, maps the stabilized olfactory sequence to the embedding space using the activation function $\mathrm{ReLU}(\cdot)$ as shown in Eq.~\eqref{eq:start_1} and Eq.~\eqref{eq:end_1}:
\begin{align}
    \mathbf{O} &= \mathrm{ReLU}(f_{\text{PLR}}(\mathbf{S}))\label{eq:start_1}\\ 
    f_{\text{PLR}}(\mathbf{S}) &= [\sin(2\pi \mathbf{W}^\top \mathbf{S}), \cos(2\pi \mathbf{W}^\top \mathbf{S})]\label{eq:end_1}
\end{align}
$\mathbf{S} \in \mathbb{R}^{N \times d_o}$ and $\mathbf{W} \in \mathbb{R}^{N \times k}$, where $N$ denotes the number of samples and $d_o$ represents the input dimensionality of the olfactory modality. Each trainable parameter $w_{i, j}$ in $\mathbf{W}$ is initialized according to $\mathcal{N}(0, \sigma^2)$. Both $\sigma$ and $k$ serve as important hyperparameters.

For the visual modality, FDEC partitions the input image $i_n \in \mathbb{R}^{H \times W \times C}$ into $L$ local patches, denoted as $\{i_{n,l}\}_{l=1}^{L}$\cite{[19]ViT}. Each patch is independently processed by the patch-level convolutional embedder (PLCE) to perform local convolutional embedding. The convolutional embedder can be formulated as a function $f_{\text{patch}}(\cdot)$ that maps the input to the visual embedding space as shown in Eq.~\eqref{eq:start_2} and Eq.~\eqref{eq:end_2}:
\begin{align}
    \mathbf{V} &= \{\mathbf{v}_n\}_{n=1}^{N}\label{eq:start_2}\\
    \mathbf{v}_n &= \{f_{\text{patch}}(i_{n,l})\}_{l=1}^{L}\label{eq:end_2}
\end{align}
where $\mathbf{v}_n$ denotes the visual embedding of the $n$-th sample $i_n$. 

Subsequently, FDEC integrates the visual embeddings $\mathbf{V}$ and the numerical olfactory embeddings $\mathbf{O}$ into a unified, structurally managed multimodal embedding set $\mathbf{X}$:
\begin{align}
    \mathbf{X} = \{\mathbf{O}, \mathbf{V}\}.\label{eq:5}
\end{align}

Within the olfactory modality, PLR performs feature enhancement on the ten-channel gas sensor signals, enabling critical variation patterns to be prominently represented. Within the visual modality, the input image $I$ is divided into $L = 9$ local patches and mapped into the embedding space, enabling the model to focus on micro-scale mold textures and local details. The resulting multimodal embedding set $\mathbf{X}$ is organized in a structured manner to ensure accessibility, allowing the visual and olfactory branches of FDRA-Net to extract their respective embeddings from a unified data structure. The FDEC procedure is presented in Algorithm \ref{FDEC_alg}.

\begin{algorithm}[h]
\caption{FDEC.}
\begin{algorithmic}
\STATE
\textbf{Require:} $\mathbf{I}=\{i_n\}_{n=1}^N$,$\mathbf{S}$\\
\textbf{Ensure:} $\mathbf{X}$\\
\hspace{0.3cm}1: $\mathbf{O} \leftarrow \mathrm{ReLU}(f_{\text{PLR}}(\mathbf{S}))$\\
\hspace{0.3cm}2: $\mathbf{V} \leftarrow \{\{f_{\text{patch}}(i_{n,l})\}_{l=1}^L\}_{n=1}^N$\\
\hspace{0.3cm}3: $\mathbf{X} \leftarrow \{\mathbf{O}, \mathbf{V}\}$\\
\textbf{return:} $\mathbf{X}$
\end{algorithmic}
\label{FDEC_alg}
\end{algorithm}

\begin{figure*}[h]
\centering
\includegraphics[width=7.0in]{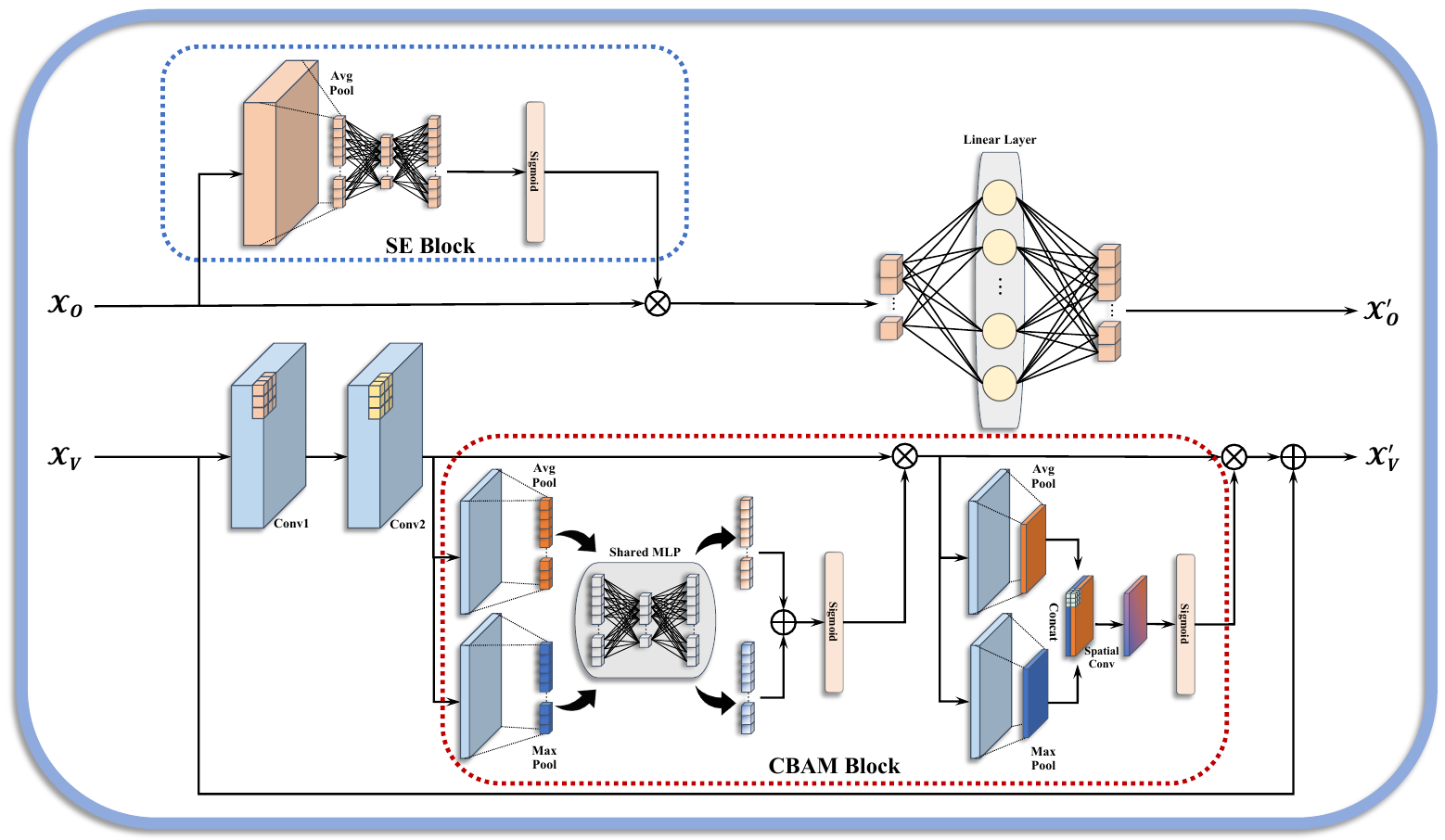}
\caption{The Structure of FDRA.}
\label{FDRA_structure}
\end{figure*}

\subsection{Multimodal Deterioration Detection Module}
Rice deterioration typically manifests as fine-grained, localized, and non-uniform abnormal patterns, posing significant challenges for multimodal detection. In the visual modality, mold spots are often sparse, small-scale, and irregularly distributed, making them susceptible to background interference and texture similarity, which hinders the stable localization of critical degradation regions\cite{[20]}. In the olfactory modality, the e-nose signals are jointly formed by the responses of multiple gas-sensitive materials, with markedly imbalanced contributions across channels, where subtle inter-channel differences play a crucial role in category discrimination\cite{[21]}. When features from the two modalities are directly fused, the model is prone to interference from redundant information and ineffective features, thereby weakening its discriminative capability. To address this issue, FDRA-Net is proposed as the core network architecture, introducing FDRA to perform modality-aware feature recalibration and enhancement, thereby improving the model’s ability to attend to discriminative visual regions and informative olfactory channels.

The FDRA-Net takes the structured multimodal embedding data constructed by the FDEC as input and separately retrieves visual embeddings and olfactory embeddings from the unified data structure, which are then fed into the corresponding visual and olfactory branches as shown in Fig. \ref{proposed_framework}. The network backbone consists of multiple stackable FDRA layers, which are used to progressively enhance and recalibrate the representations of the two modalities. Subsequently, linear projection layers are employed to organize the enhanced features in terms of dimensionality, and the resulting representations are fed into a classifier to perform three-class decision making, producing predictions for the expired, moldy, and normal categories. By leveraging the FDRA to selectively strengthen discriminative information in both visual and olfactory features at the feature level, this module enables the two modalities to collaboratively and complementarily support the rice deterioration detection task.

The FDRA consists of a squeeze-and-excitation (SE)-based channel recalibration unit\cite{[22]SE} on the olfactory side and a convolutional block attention module (CBAM)\cite{[23]CBAM} on the visual side, which are designed to differentially enhance the discriminative information of the two modalities as shown in Fig.~\ref{FDRA_structure}. The embedded feature data are retrieved from the multimodal embedded-feature dataset constructed by the FDEC. On the olfactory side, the SE generates channel-wise weights to perform recalibration, followed by a linear mapping to produce the olfactory output of the current layer. On the visual side, the CBAM is employed to enhance responses related to sparse, small-scale, and spatially irregular mold patterns. The processing of a single FDRA layer is formalized as shown in Eqs.~\eqref{eq:start_3}--\eqref{eq:end_3}:

\begin{align}
\mathbf{a}_o &= \sigma\!\left(\mathrm{MLP}\big(\mathrm{AvgPool}(\mathbf{x}_o)\big)\right)\label{eq:start_3}, \\
\tilde{\mathbf{x}}_o &= \mathbf{a}_o \odot \mathbf{x}_o, \\
\mathbf{h}_o &= g_o(\tilde{\mathbf{x}}_o) = W_o\,\tilde{\mathbf{x}}_o + \mathbf{b}_o, \\[2pt]
\mathbf{r}_v &= \mathcal{F}(\mathbf{x}_v;\theta_v), \\
\tilde{\mathbf{r}}_v &= \mathrm{CBAM}(\mathbf{r}_v), \\
\mathbf{h}_v &= \tilde{\mathbf{r}}_v + \mathcal{D}(\mathbf{x}_v).\label{eq:end_3}
\end{align}

In FDRA-Net, multiple FDRA layers are stacked to progressively enhance the representations of the two modalities. Let the input of the $m$-th layer be $(\mathbf{x}_o^{(m-1)}, \mathbf{x}_v^{(m-1)})$, and the corresponding output be $(\mathbf{h}_o^{(m)}, \mathbf{h}_v^{(m)})$, which serves as the input to the next layer:

\begin{align}
\left(\mathbf{x}_o^{(m)}, \mathbf{x}_v^{(m)}\right)
=
\left(\mathbf{h}_o^{(m)}, \mathbf{h}_v^{(m)}\right),
\qquad m=1,2,\dots,M
\end{align}

where $(\mathbf{h}_o^{(m)}, \mathbf{h}_v^{(m)})$ are computed by the FDRA layer defined above. After stacking $M$ layers, FDRA-Net applies linear layers to vectorize and map the outputs of the two modalities, which are then fed into a classification head to obtain the final predictions:

\begin{align}
\mathbf{z}_o &= \mathrm{Linear}_o\!\left(\psi_o(\mathbf{x}_o^{(M)})\right), \\
\mathbf{z}_v &= \mathrm{Linear}_v\!\left(\psi_v(\mathbf{x}_v^{(M)})\right), \\
\hat{\mathbf{y}} &= \mathrm{Softmax}\!\left(
f_{\mathrm{cls}}\Big([\mathbf{z}_v \ \| \ \mathbf{z}_o]\Big)
\right).
\end{align}

The $\mathbf{x}_o$ and $\mathbf{x}_v$ denote the input features of the olfactory and visual modalities at a FDRA layer, respectively. $\mathrm{AvgPool}(\cdot)$, $\mathrm{MLP}(\cdot)$, and $\sigma(\cdot)$ represent global average pooling, a two-layer perceptron, and the Sigmoid activation function, respectively, and $\odot$ denotes element-wise multiplication. $g_o(\cdot)$ denotes the linear mapping on the olfactory side. On the visual side, $\mathcal{F}(\cdot;\theta_v)$ and $\mathcal{D}(\cdot)$ denote the residual branch and the shortcut branch, respectively, and $\mathrm{CBAM}(\cdot)$ is inserted between the residual branch output and the residual summation. $\psi_o(\cdot)$ and $\psi_v(\cdot)$ denote vectorization operations, $\mathrm{Linear}_o(\cdot)$ and $\mathrm{Linear}_v(\cdot)$ denote linear layers, $[\cdot \| \cdot]$ denotes feature concatenation, $f_{\mathrm{cls}}(\cdot)$ denotes the classification head, and $\hat{\mathbf{y}}$ represents the predicted probability distribution over the three classes. The procedure is presented in Algorithm \ref{FDRA_Net_alg}.

\begin{algorithm}[h]
\caption{FDRA-Net.}
\begin{algorithmic}
\STATE
\textbf{Require:} $\mathbf{X}=\{\mathbf{O},\mathbf{V}\}$, $M$ \\
\textbf{Ensure:} $\hat{\mathbf{y}}$ \\
\hspace{0.3cm}1: $(\mathbf{x}_o^{(0)}, \mathbf{x}_v^{(0)}) \leftarrow (\mathbf{O}, \mathbf{V})$ \\
\hspace{0.3cm}2: \textbf{for} $m = 1$ \textbf{to} $M$ \textbf{do} \\
\hspace{0.3cm}3: \hspace{0.3cm} $\mathbf{a}_o^{(m)} \leftarrow \sigma\!\left(\mathrm{MLP}\big(\mathrm{AvgPool}(\mathbf{x}_o^{(m-1)})\big)\right)$ \\
\hspace{0.3cm}4: \hspace{0.3cm}$\mathbf{h}_o^{(m)} \leftarrow g_o\!\left(\mathbf{a}_o^{(m)} \odot \mathbf{x}_o^{(m-1)}\right)$ \\
\hspace{0.3cm}5: \hspace{0.3cm}$\mathbf{r}_v^{(m)} \leftarrow \mathcal{F}(\mathbf{x}_v^{(m-1)};\theta_v)$ \\
\hspace{0.3cm}6: \hspace{0.3cm}$\mathbf{h}_v^{(m)} \leftarrow \mathrm{CBAM}(\mathbf{r}_v^{(m)}) + \mathcal{D}(\mathbf{x}_v^{(m-1)})$ \\
\hspace{0.3cm}7: \hspace{0.3cm}$(\mathbf{x}_o^{(m)}, \mathbf{x}_v^{(m)}) \leftarrow (\mathbf{h}_o^{(m)}, \mathbf{h}_v^{(m)})$ \\
\hspace{0.3cm}8: \textbf{end for} \\
\hspace{0.3cm}9: $\mathbf{z}_o \leftarrow \mathrm{Linear}_o\!\left(\psi_o(\mathbf{x}_o^{(M)})\right)$ \\
\hspace{0.15cm}10: \hspace{0.15cm}$\mathbf{z}_v \leftarrow \mathrm{Linear}_v\!\left(\psi_v(\mathbf{x}_v^{(M)})\right)$ \\
\hspace{0.15cm}11: \hspace{0.15cm}$\hat{\mathbf{y}} \leftarrow \mathrm{Softmax}\!\left(f_{\mathrm{cls}}\big([\mathbf{z}_v \| \mathbf{z}_o]\big)\right)$ \\
\textbf{return:} $\hat{\mathbf{y}}$
\end{algorithmic}
\label{FDRA_Net_alg}
\end{algorithm}


\section{Experiments}

The experiments were conducted on a dataset collected by the electronic nose and RGB camera. The proposed method was evaluated in terms of its ability to capture and represent critical sample features, its sensitivity to anomalies, and its classification accuracy. Python 3.9.18 and the open-source deep learning framework PyTorch 2.6.0 were employed. The experiments were performed on a GPU-based cloud server equipped with four NVIDIA RTX A6000 GPUs and an Intel Xeon Gold 6330 CPU.

\subsection{Dataset Preparation}
The rice samples in this study were categorized into three quality levels: normal, expired, and moldy. Each sample completed the data collection procedure described above, and the samples originated from nine non-consecutive collection days with a balanced distribution. The FDEC was employed to reconstruct a multimodal embedded-feature dataset.

To prevent data leakage and ensure the reliability of model evaluation, the dataset was partitioned based on collection days. The 7,200 samples obtained from the first eight collection days were used as the training set, within which the validation set was further split, and 900 samples obtained from the ninth collection day were used as the test set. It should be noted that the temporal interval between the eighth and ninth collection days was significantly longer than the intervals among the first eight collection days.



\begin{figure}[h]
\centering
\includegraphics[width=3.5in]{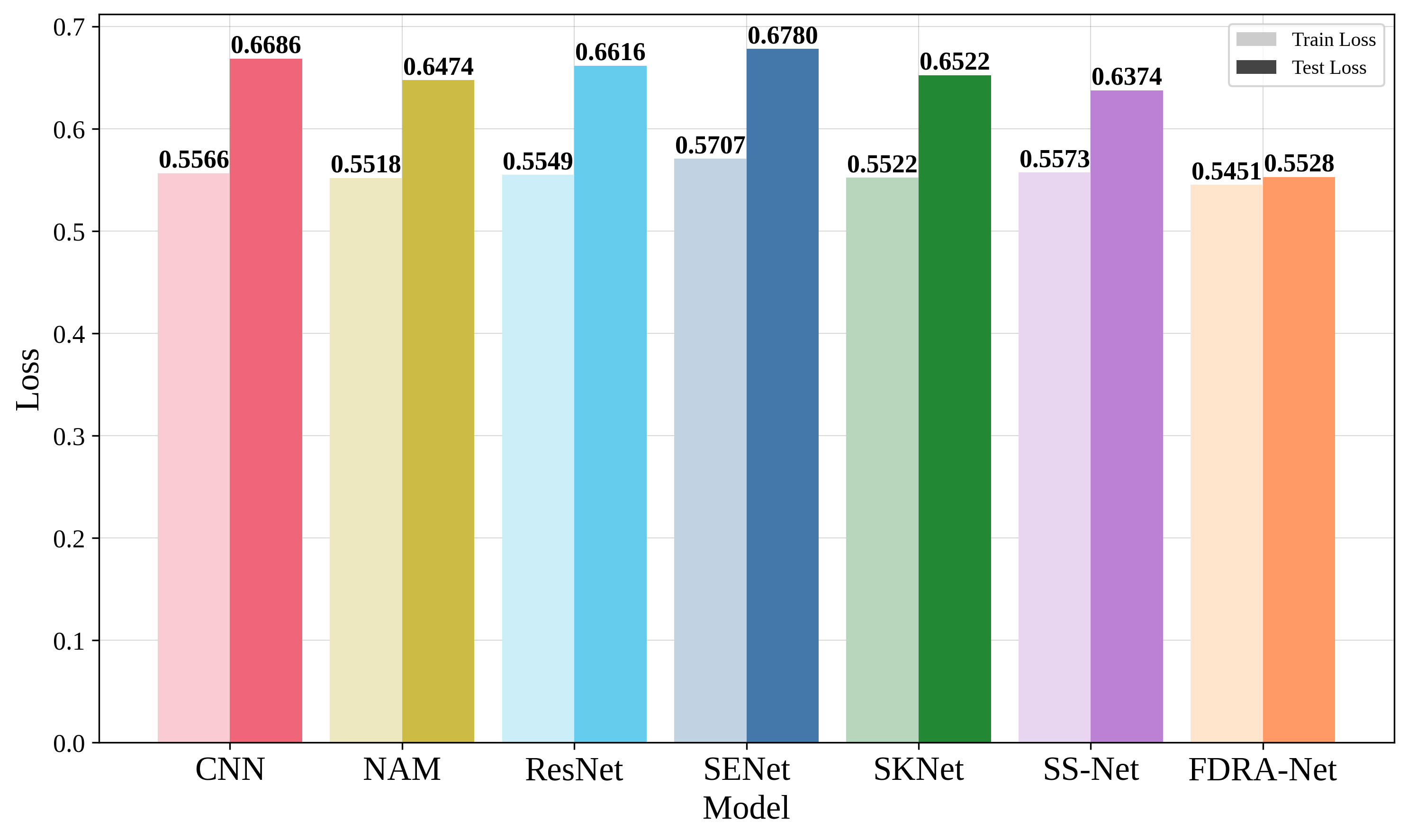}
\caption{Comparison of training loss and testing loss across different models.}
\label{fig_train_val_loss}
\end{figure}


\subsection{Offline Experiments}
During the training process, cross-entropy was selected as the loss function, and Adam was adopted as the optimizer. The experiments were conducted to evaluate the proposed method.

\subsubsection{Comparative Experiments with SOTA} To verify the effectiveness and performance advantages of the proposed FDRA-Net, the experiments were conducted. Baseline models included the convolution-based models SS-Net \cite{[24]SS-Net}, CNN \cite{[25]orgin_CNN}, ResNet\cite{[26]ResNet}, SENet\cite{[22]SE}, SKNet\cite{[27]SKNet}, and NAM\cite{[28]NAM}. In these models, SS-Net represented the most advanced model for rice detection, while the other models were commonly used in rice detection. FDRA-Net used the data that were processed by FDEC, while the other models processed the data according to the work of Shi et al.\cite{[24]SS-Net}. To ensure more accurate experimental results, grid search and Bayesian optimization were used for hyperparameter tuning. FDRA-Net and the baseline models had their core layers set to four, were trained and tested with the batch size set to 256 and the number of epochs set to 100. 

The evaluation metrics of the experiments were presented in Table \ref{table_baseline}. The experimental results showed that the baseline models had smaller sizes and shorter inference latency. The expired category was more difficult to identify and was easily confused with the normal category; therefore, several baseline models exhibited lower Recall in this category. In contrast, FDRA-Net achieved a Recall of 99.67\% and an F1 Score of 99.83\% for the expired category, indicating that FDRA-Net had stronger robustness in capturing fine-grained degradation features. Compared with the baseline models, FDRA-Net exhibited lower training and testing losses, and the difference between the two was small, suggesting that FDRA-Net achieved better fitting stability on samples from the same distribution as shown in Fig. \ref{fig_train_val_loss}.

Although the architecture of FDRA-Net was more complex and its model size was larger, resulting in longer inference latency, its accuracy improved significantly. This indicated that FDRA-Net achieved substantial performance gains at the cost of a moderate increase in computational overhead. In addition, although the size and latency of FDRA-Net were greater than those of the baseline models, its model size was 48.73 MB and its inference latency was 9.483 ms, which still satisfied the requirements for deployment on industrial or embedded devices and for real-time detection.

FDRA-Net achieved an 8.67\% improvement in overall accuracy compared with SS-Net and an average accuracy improvement of 11.51\% over other baseline models. These quantitative results indicated that FDRA-Net significantly outperformed existing methods across multiple evaluation metrics. 

\subsubsection{Interpretability Analysis of the FDRA Module} To further verify the role of the FDRA module in the model’s decision-making process from a qualitative perspective, the experiments were conducted. The Grad-CAM was employed on the visual branch to compare the complete FDRA-Net with the model in which the FDRA module was removed.

Table \ref{tabular_gradcam} presented the Grad-CAM results of the two configurations on the same sample. The visualizations showed that, without the FDRA module, the activation regions were relatively scattered, and the high-response areas exhibited weak correspondence with the actual mold-related surface textures. In contrast, the complete FDRA-Net produced more concentrated activation distributions that highlighted the localized degradation structures associated with mold, demonstrating stronger semantic consistency. This phenomenon indicated that the FDRA module guided the visual branch to focus on more discriminative fine-grained degradation regions, thereby enhancing the model’s classification capability.

FDRA-Net exhibited remarkable performance advantages in this experiment. Subsequent experiments further analyzed the roles of its core processing method and module, validating the specific contributions of FDEC and FDRA to the performance improvement of the proposed framework.

\begin{table*}[t!]
\centering
\caption{Comparison of Models}
\begin{tabular}{lc|cccc|cc}
\toprule               
Models & Accuracy(\%)  & Category & Precision(\%) & Recall(\%) & F1-Score(\%) & Size(MB) & Latency(ms)\\  
\midrule              
& & Expired & 88.71 & 73.33 & 80.29 & &  \\ 
CNN & 87.78 & Moldy & 92.43 & 97.67 & 94.98 & 4.14 & \textbf{0.461}\\
& & Normal & 82.69 & 92.33 & 87.24 & &\\
\midrule

& & Expired & 90.15 & 79.33 & 84.40 & &  \\ 
NAM & 89.89 & Moldy & 96.64 & 96.00 & 96.32 & 4.15 & 0.677\\
& & Normal & 83.73 & 94.33 & 88.71 & &\\
\midrule

& & Expired & 87.08 & 78.67 & 82.66 & &  \\ 
ResNet & 88.33 & Moldy & 90.42 & 94.33 & 92.33 & 4.32 & 0.665\\
& & Normal & 87.34 & 92.00 & 89.61 & &\\
\midrule

& & Expired & 90.42 & 72.33 & 80.37 & &  \\ 
SENet & 86.78 & Moldy & 89.14 & 93.00 & 91.03 & \textbf{4.08} & 0.846\\
& & Normal & 82.13 & 95.00 & 88.10 & &\\
\midrule

& & Expired & 86.69 & 80.33 & 83.39 & &  \\ 
SKNet & 89.22 & Moldy & 86.35 & 97.00 & 91.37 & 4.67 & 1.641\\
& & Normal & 95.09 & 90.33 & 92.65 & &\\
\midrule

& & Expired & 82.42 & 95.33 & 88.41 & &  \\ 
SS-Net & 91.22 & Moldy & 98.32 & 97.67 & 97.99 & 4.38 & 2.124\\
& & Normal & 94.90 & 80.67 & 87.21 & &\\
\midrule

& & Expired & \textbf{100.00} & \textbf{99.67} & \textbf{99.83} & &  \\ 
\textbf{FDRA-Net} & \textbf{99.89} & Moldy & \textbf{100.00} & \textbf{100.00} & \textbf{100.00} & 48.73\textsuperscript{\textbf{*}} & 9.483\textsuperscript{\textbf{*}}\\
& & Normal & \textbf{99.67} & \textbf{100.00} & \textbf{99.83} & &\\

\bottomrule           
\end{tabular}

\vspace{2mm}
\footnotesize
\justifying
\textbf{Note:}
\textsuperscript{\textbf{*}} marks the comparatively higher size and latency of FDRA-Net; however, its latency (9.483 ms) is negligible relative to spectroscopic acquisition time\cite{[29]} and is unlikely to be a practical bottleneck.

\label{table_baseline}
\end{table*}


\begin{table}[t]
\centering
\caption{Qualitative Visualization of the Effectiveness of the FDRA Module}
\begin{tabular}{c | c | c c}
\toprule
\makecell{Input\\Samples} &
\makecell{FDRA} &
\makecell{Grad-CAM\\Heatmap} &
\makecell{Overlay\\Visualization} \\
\midrule

\multirow{2}{*}[2.35em]{%
    \includegraphics[height=2.0cm]{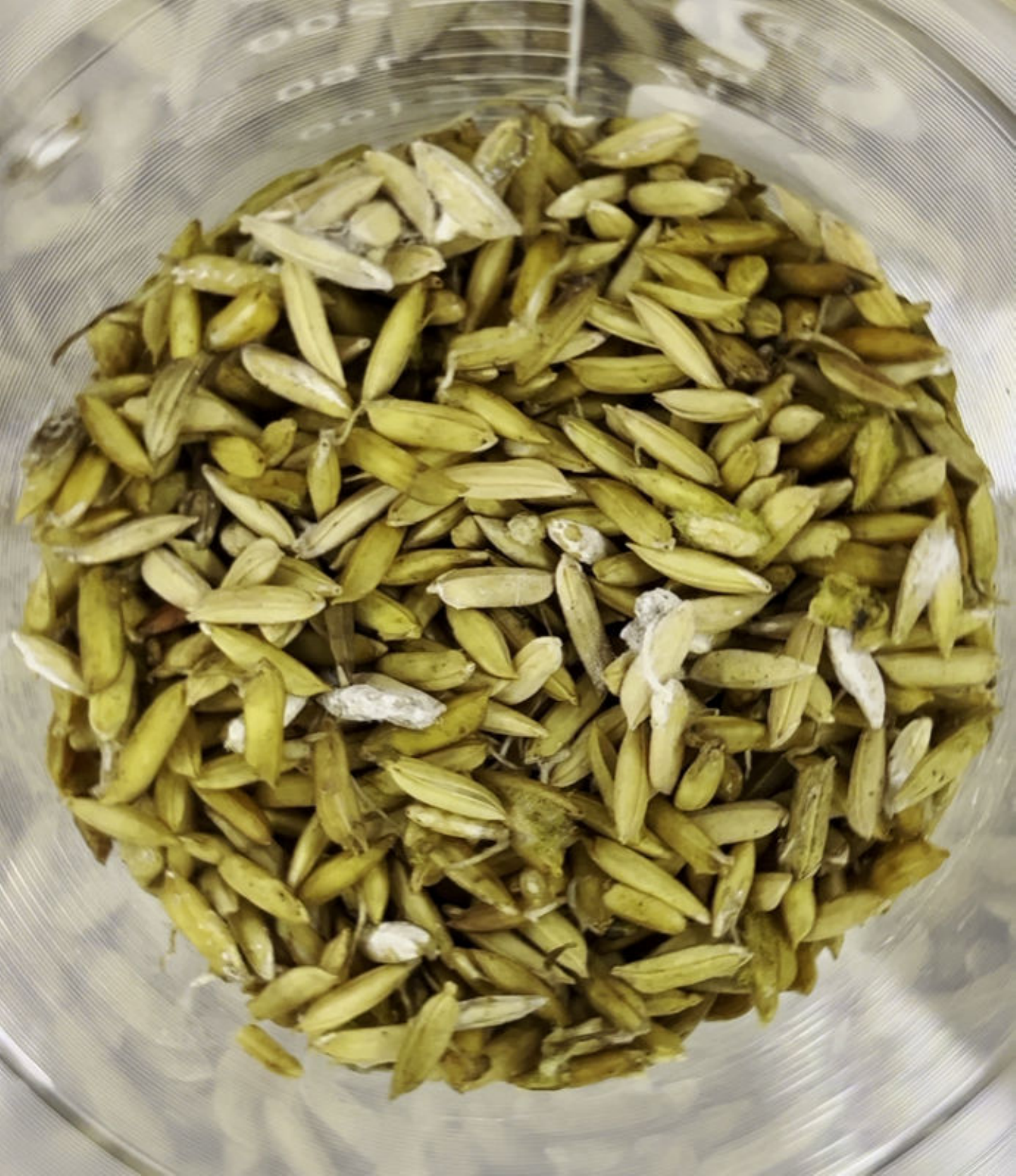}
}

& \raisebox{3.5em}{---}
& \includegraphics[height=2.0cm]{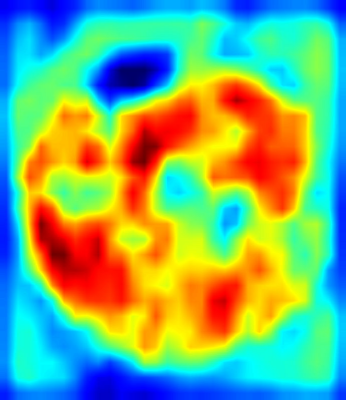}
& \includegraphics[height=2.0cm]{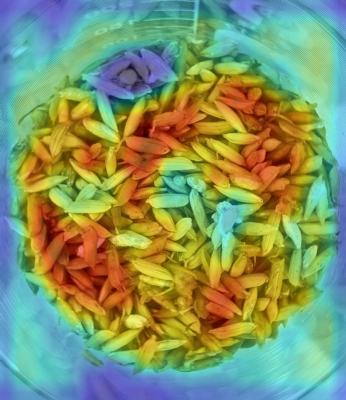}
\\

& \raisebox{3.5em}{\checkmark}
& \includegraphics[height=2.0cm]{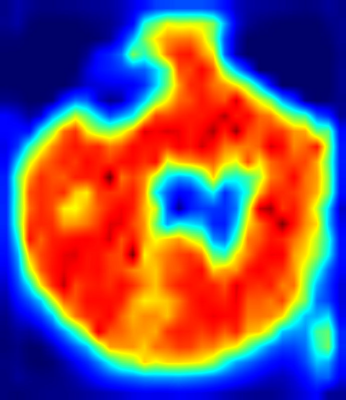}
& \includegraphics[height=2.0cm]{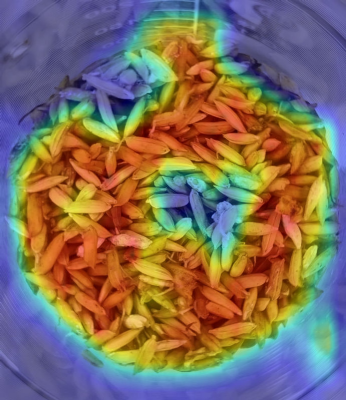}
\\

\midrule

\multirow{2}{*}[2.35em]{%
    \includegraphics[height=2.0cm]{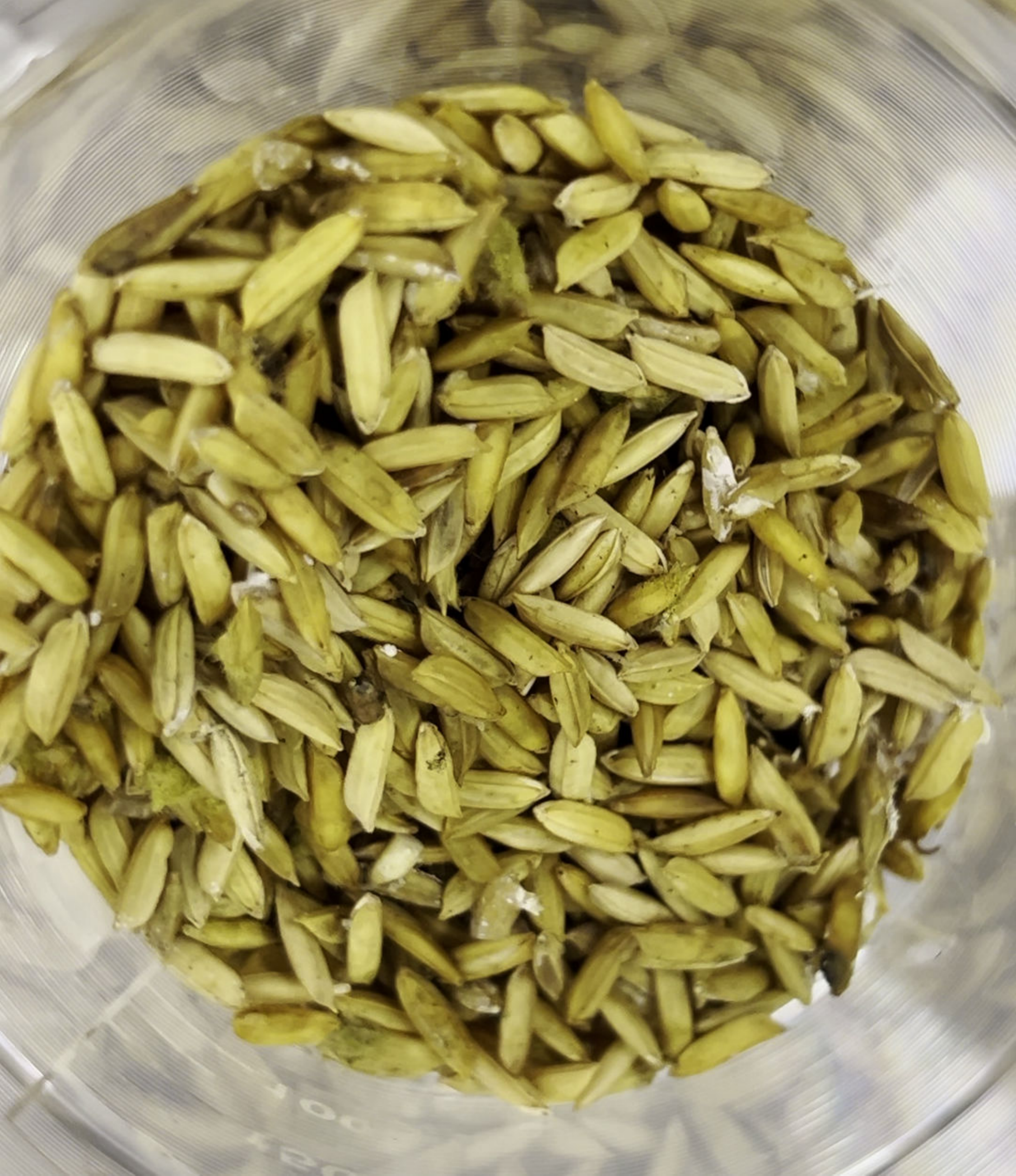}
}

& \raisebox{3.5em}{---}
& \includegraphics[height=2.0cm]{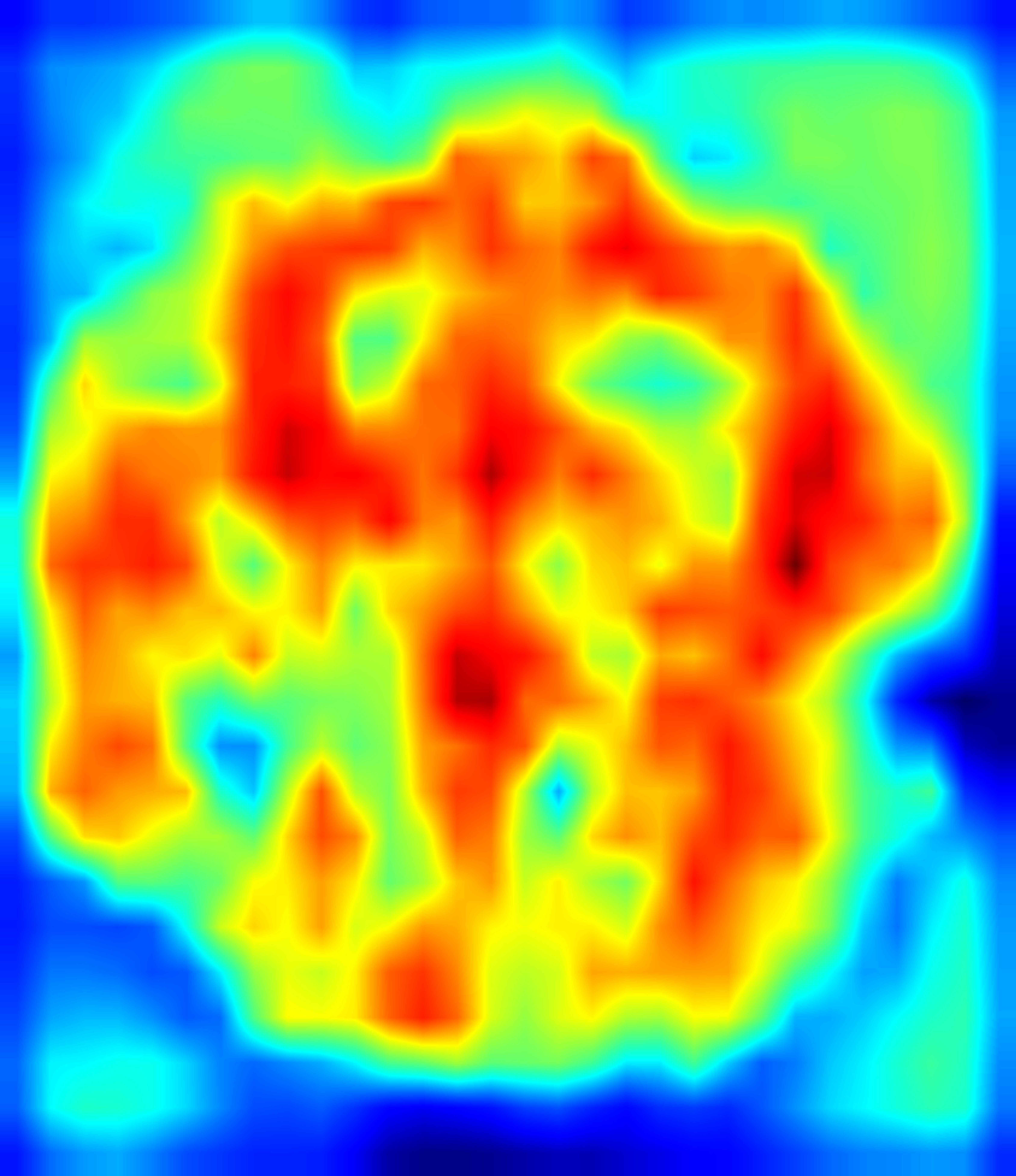}
& \includegraphics[height=2.0cm]{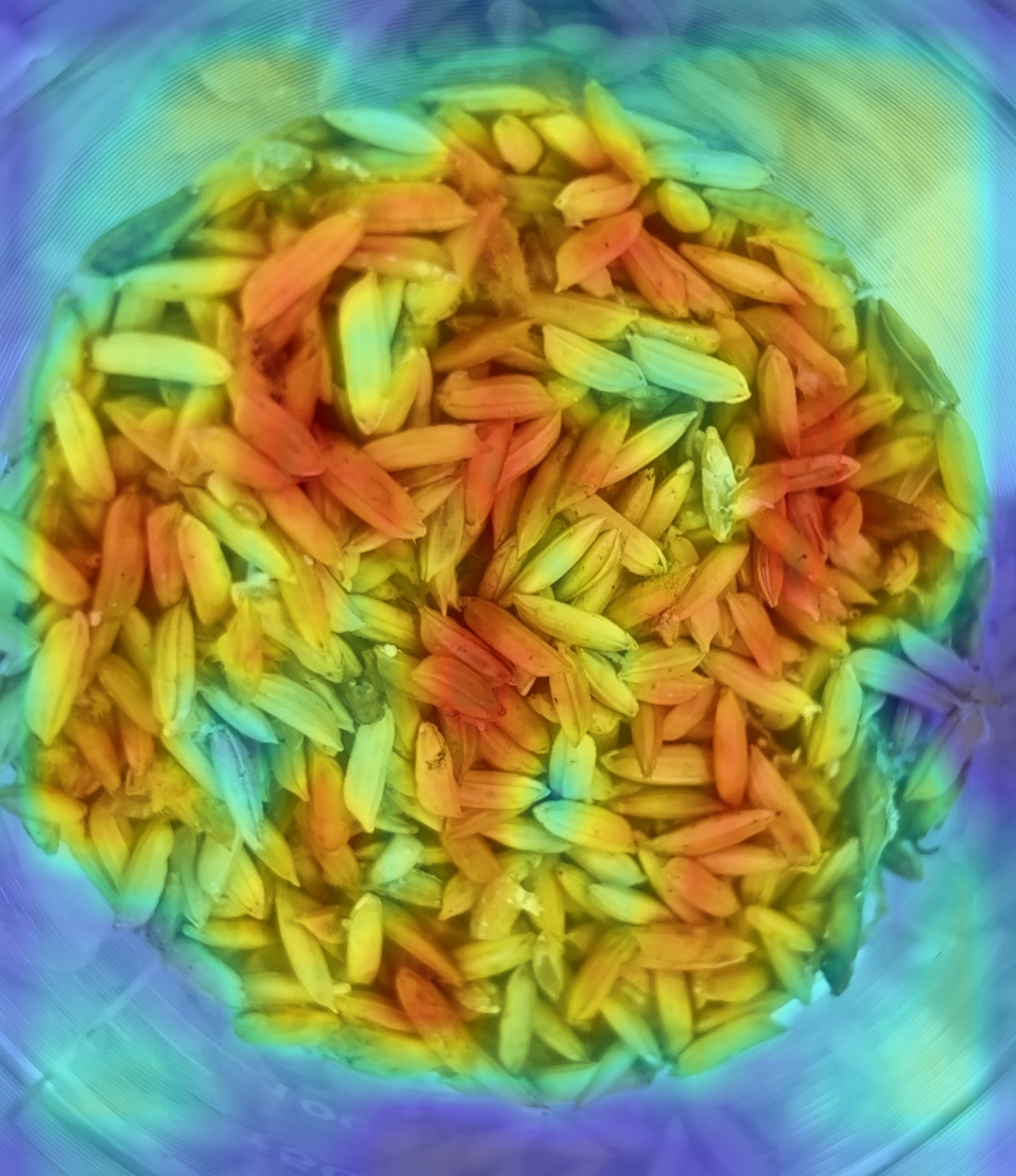}
\\

& \raisebox{3.5em}{\checkmark}
& \includegraphics[height=2.0cm]{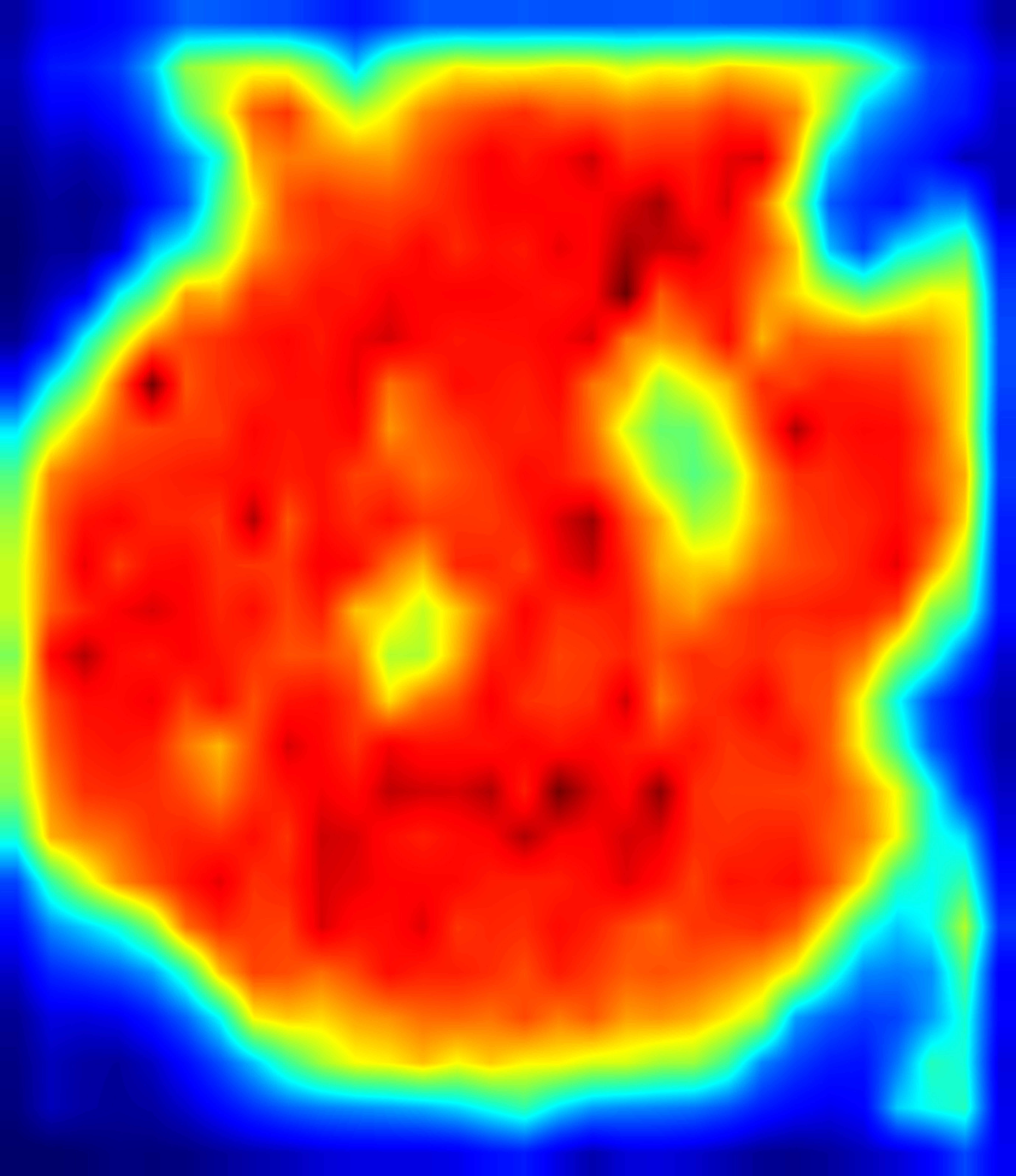}
& \includegraphics[height=2.0cm]{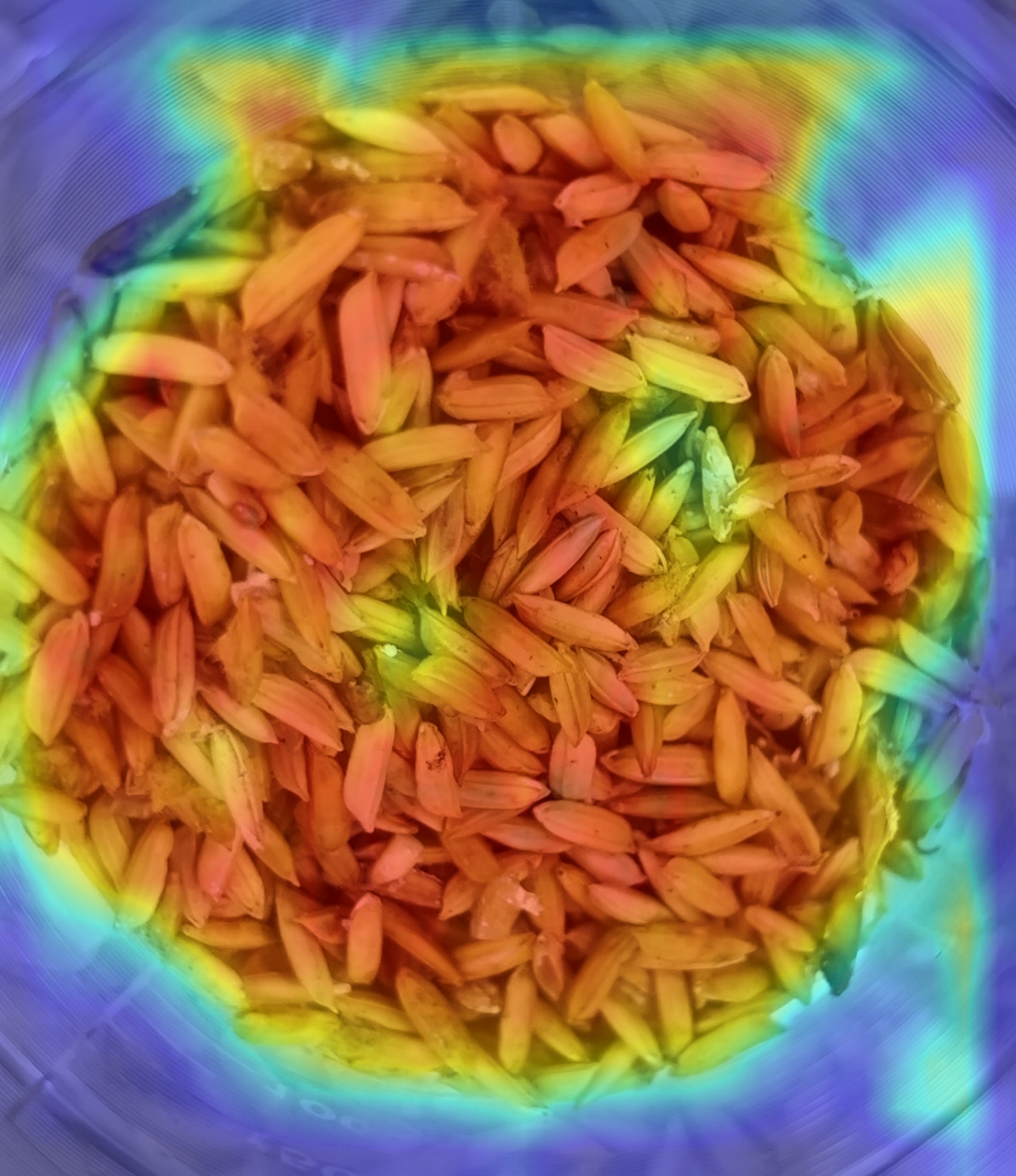}
\\

\midrule

\multirow{2}{*}[2.35em]{%
    \includegraphics[height=2.0cm]{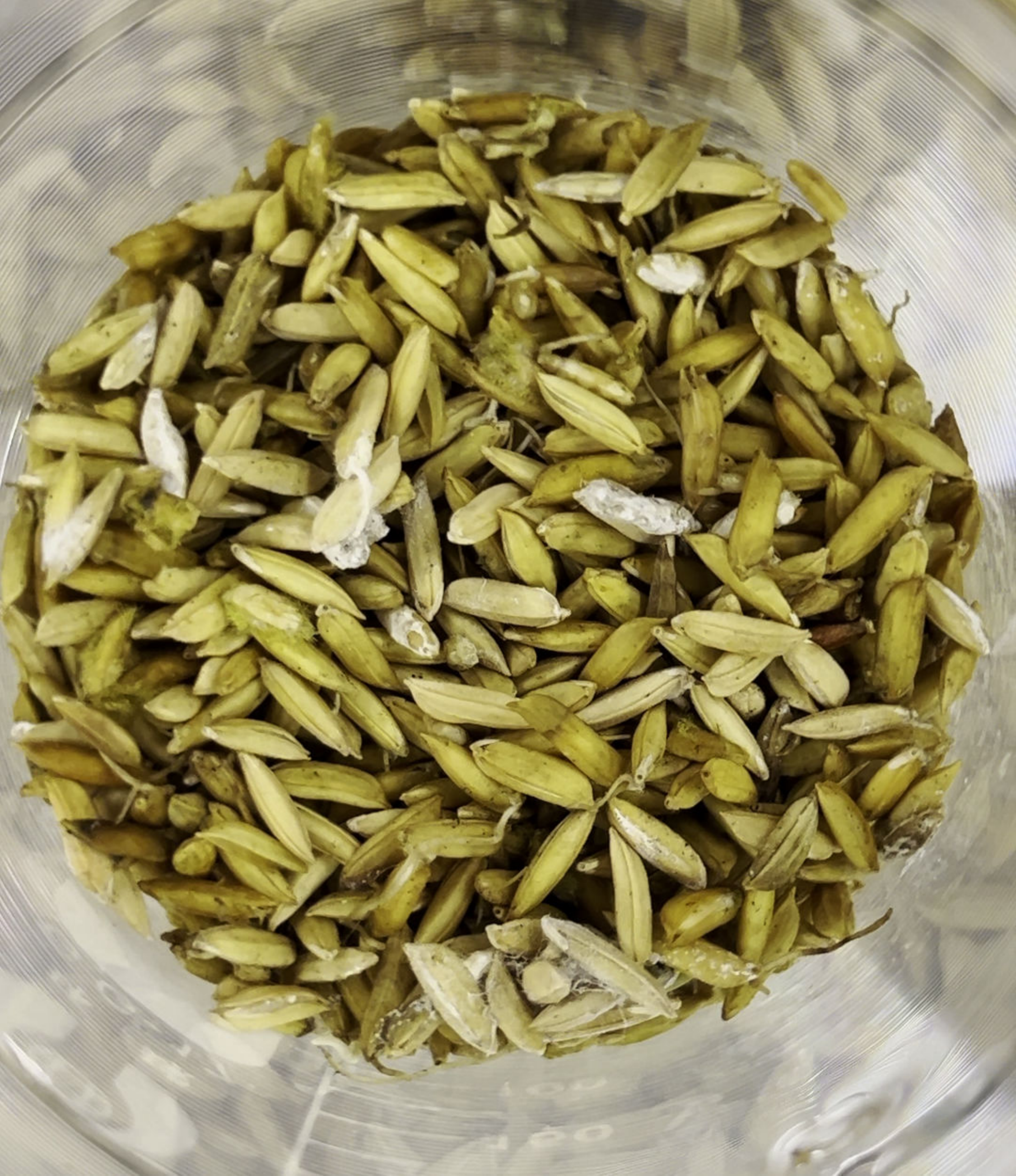}
}

& \raisebox{3.5em}{---}
& \includegraphics[height=2.0cm]{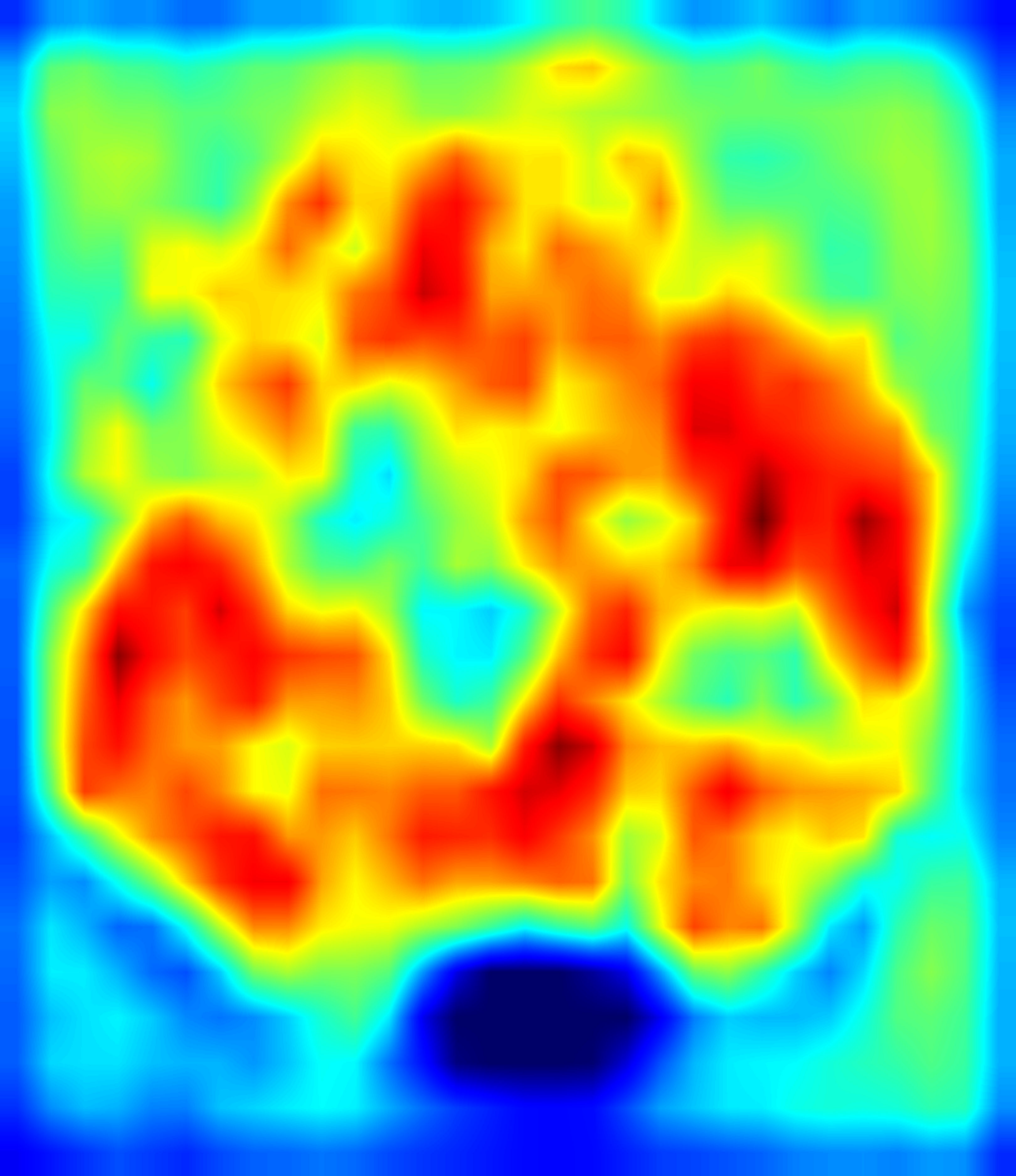}
& \includegraphics[height=2.0cm]{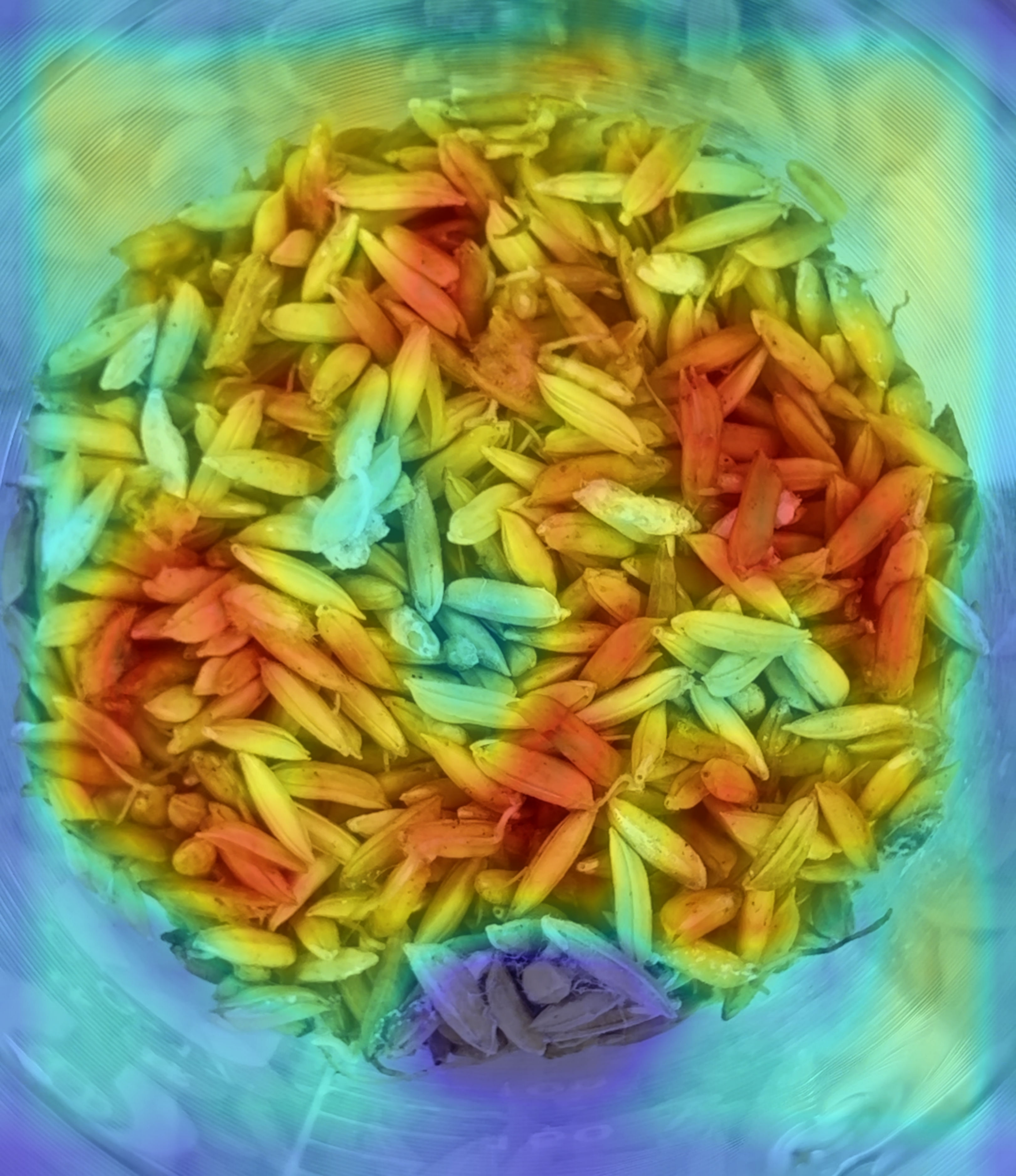}
\\

& \raisebox{3.5em}{\checkmark}
& \includegraphics[height=2.0cm]{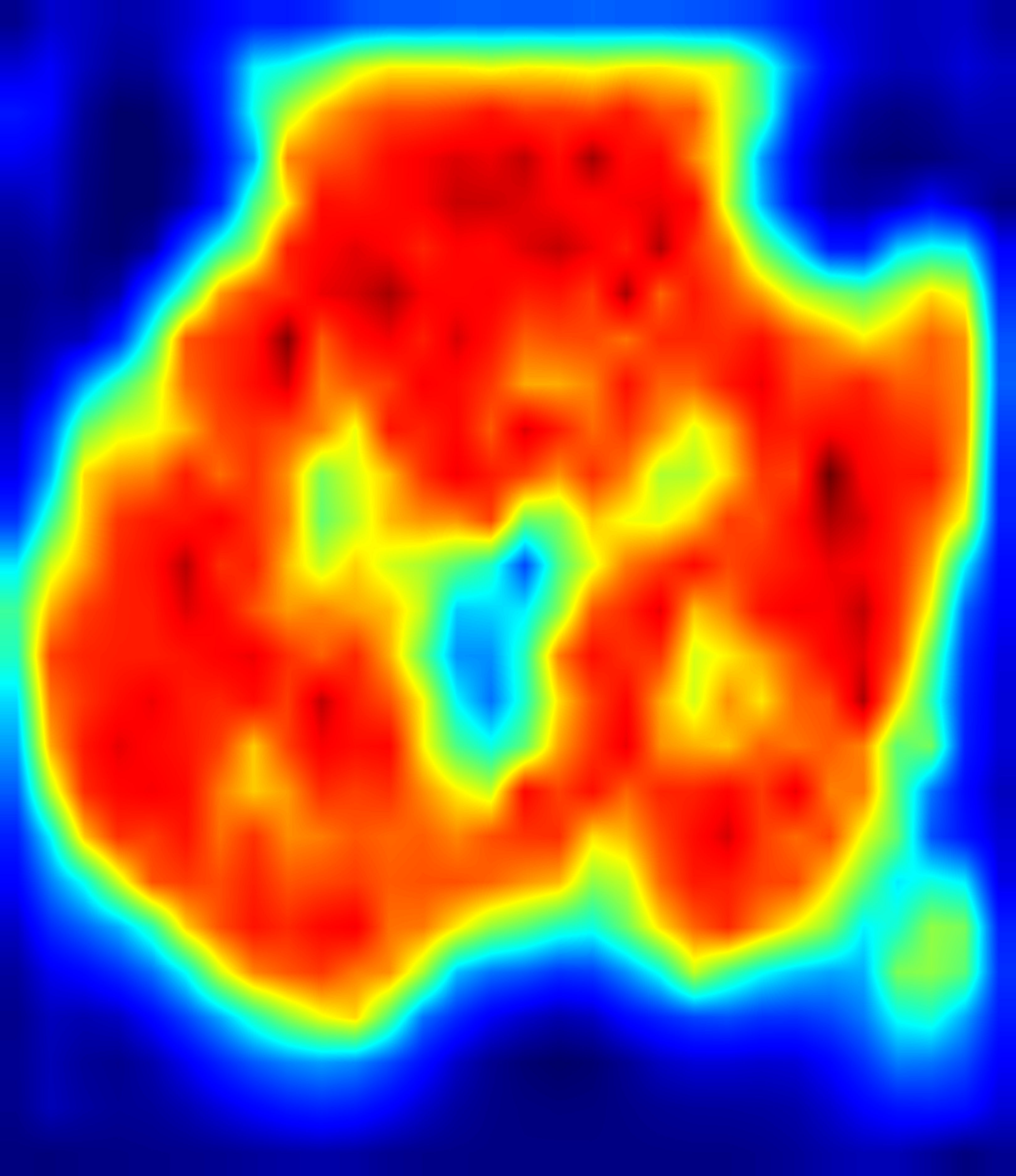}
& \includegraphics[height=2.0cm]{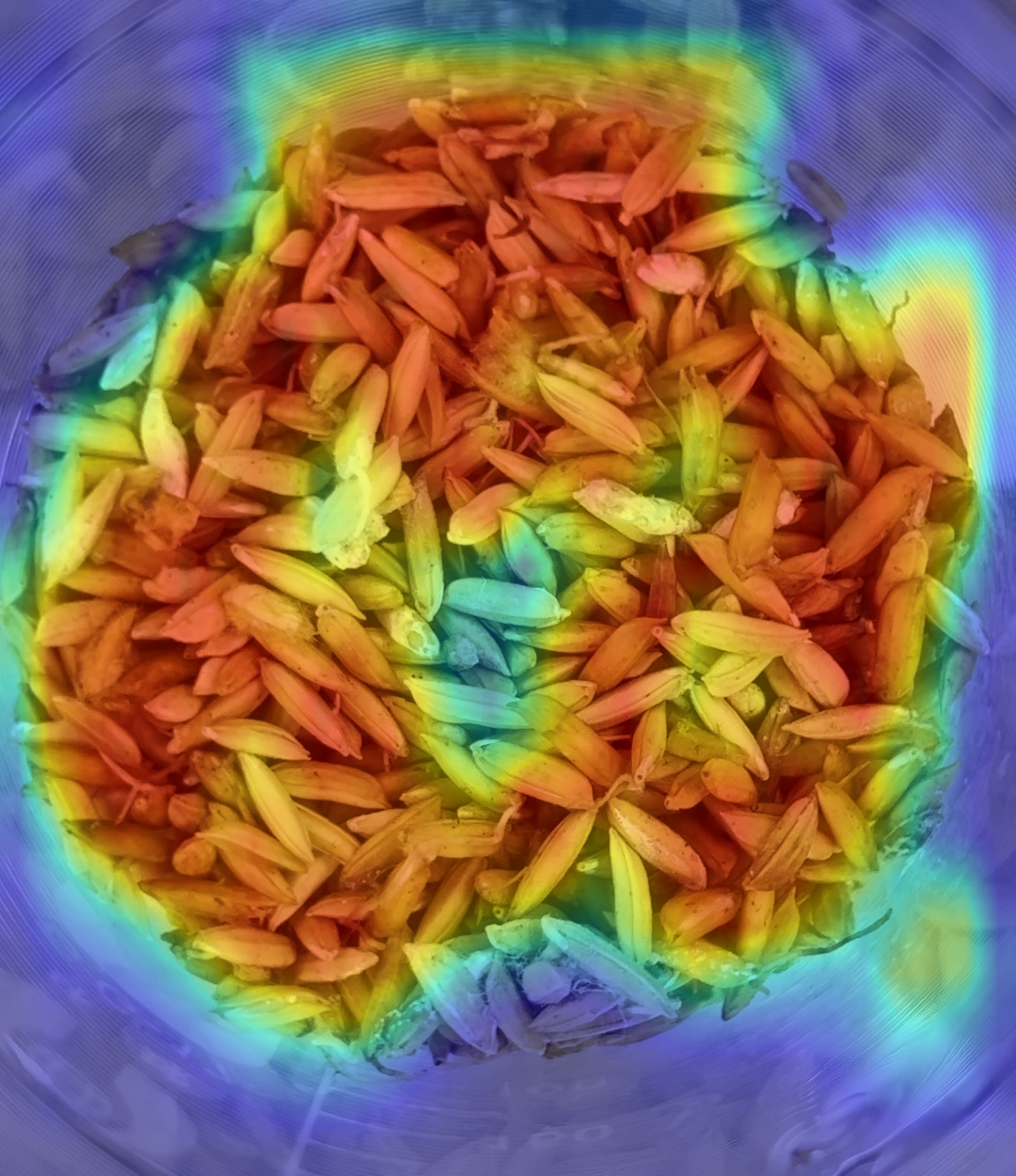}
\\

\bottomrule
\end{tabular}
\label{tabular_gradcam}
\end{table}


\begin{table*}[h]
\centering
\caption{Ablation Results of The Proposed Method}
\begin{tabular}{@{}ccc|cc|ccccc@{}}
\toprule
\multicolumn{1}{c|}{FDEC} & \multicolumn{2}{c|}{FDRA Submodules} 
& Accuracy(\%) & $\Delta$Acc(\%) & Category & Precision(\%) & Recall(\%) & F1-Score(\%)\\
\cmidrule(lr){2-3}
\multicolumn{1}{c|}{} & SE & CBAM & & & & & & \\
\midrule

\multirow{3}{*}{\checkmark} & \multirow{3}{*}{\checkmark} & \multirow{3}{*}{\checkmark} 
& \multirow{3}{*}{99.89} & \multirow{3}{*}{---} & Expired & 100.00 & 99.67 & 99.83\\
& & & & & Moldy & 100.00 & 100.00 & 100.00\\
& & & & & Normal & 99.67 & 100.00 & 99.83\\
\midrule

\multirow{3}{*}{---} & \multirow{3}{*}{\checkmark} & \multirow{3}{*}{\checkmark}
& \multirow{3}{*}{85.44} & \multirow{3}{*}{-14.45} & Expired & 81.48 & 80.67 & 81.07\\
& & & & & Moldy & 88.68 & 78.33 & 83.19\\
& & & & & Normal & 86.39 & 97.33 & 91.54\\
\midrule

\multirow{3}{*}{\checkmark} & \multirow{3}{*}{---} & \multirow{3}{*}{\checkmark}
& \multirow{3}{*}{94.22} & \multirow{3}{*}{-5.67} & Expired & 91.11 & 95.67 & 93.33\\
& & & & & Moldy & 95.85 & 92.33 & 94.06\\
& & & & & Normal & 95.95 & 94.67 & 95.30\\
\midrule

\multirow{3}{*}{\checkmark} & \multirow{3}{*}{\checkmark} & \multirow{3}{*}{---}
& \multirow{3}{*}{94.89} & \multirow{3}{*}{-5.00} & Expired & 89.39 & 98.33 & 93.65\\
& & & & & Moldy & 97.79 & 88.67 & 93.01\\
& & & & & Normal & 98.32 & 97.67 & 97.99\\
\midrule

\multirow{3}{*}{\checkmark} & \multirow{3}{*}{---} & \multirow{3}{*}{---}
& \multirow{3}{*}{90.67} & \multirow{3}{*}{-9.22} & Expired & 83.09 & 95.32 & 88.79\\
& & & & & Moldy & 94.40 & 84.05 & 88.93\\
& & & & & Normal & 96.19 & 92.67 & 94.40\\
\bottomrule
\end{tabular}
\label{table_ablation}
\end{table*}

\subsection{Ablation Experiments} 
To verify the effects of various components in the FDRA-Net, ablation experiments were conducted. The PLR and PLCE were applied to construct FDEC, aiming to enhance the representational capability of the e-nose data and RGB image data. FDRA consisted of convolutional layers integrated with CBAM and linear layers integrated with SE, designed to focus on the signal differences of the e-nose’s gas-sensitive materials when measuring different samples and to enhance sensitivity to the fine-grained rice deterioration.

\begin{table*}[!t]
\centering
\caption{Comparison of Field Sample Predictions Across Models}
\begin{tabular}{lc|ccccccc}
\toprule
\multicolumn{1}{l|}{\multirow{2}{*}{Samples}} & \multirow{2}{*}{Labels} & \multicolumn{7}{c}{Models}                             \\ \cline{3-9} 
\multicolumn{1}{l|}{}                        &                             & CNN & NAM & ResNet & SENet & SKNet & SS-Net & \textbf{FDRA-Net} \\ \midrule
\multicolumn{1}{l|}{No.1}                    &             0               &  0  &  0  &    0   &   0   &   0   &   0    &     0    \\
\multicolumn{1}{l|}{No.2}                    &             1               &  1  &  \fbox{0}  &    1   &   1   &   1   &   1    &     1    \\
\multicolumn{1}{l|}{No.3}                    &             0               &  0  &  0  &    0   &   0   &   0   &   0    &     0    \\
\multicolumn{1}{l|}{No.4}                    &             2               &  2  &  2  &    \fbox{0}   &   2   &   2   &   2    &     2    \\
\multicolumn{1}{l|}{No.5}                    &             0               &  \fbox{1}  &  \fbox{2}  &    0   &   \fbox{2}   &   0   &   0    &     \fbox{2}    \\
\multicolumn{1}{l|}{No.6}                    &             1               &  1  &  1  &    \fbox{0}   &   1   &   \fbox{0}   &   1    &     1    \\
\multicolumn{1}{l|}{No.7}                    &             0               &  0  &  0  &    0   &   0   &   0   &   0    &     0    \\
\multicolumn{1}{l|}{No.8}                    &             2               &  \fbox{0}  &  2  &    2   &   2   &   \fbox{0}   &   2    &     2    \\
\multicolumn{1}{l|}{No.9}                    &             0               &  0  &  0  &    0   &   0   &   0   &   0    &     0    \\
\multicolumn{1}{l|}{No.10}                   &             1               &  \fbox{0}  &  1  &    \fbox{0}   &   1   &   1   &   1    &     1    \\
\multicolumn{1}{l|}{No.11}                   &             2               &  2  &  2  &    2   &   \fbox{0}   &   \fbox{0}   &   \fbox{0}    &     2    \\
\multicolumn{1}{l|}{No.12}                   &             0               &  0  &  0  &    0   &   0   &   0   &   \fbox{1}    &     0    \\
\multicolumn{1}{l|}{No.13}                   &             1               &  \fbox{0}  &  1  &    1   &   1   &   1   &   1    &     1    \\
\multicolumn{1}{l|}{No.14}                   &             0               &  \fbox{2}  &  0  &    \fbox{2}   &   0   &   0   &   0    &     0    \\
\multicolumn{1}{l|}{No.15}                   &             2               &  2  &  \fbox{0}  &    2   &   2   &   2   &   2    &     2    \\ \midrule
\multicolumn{2}{c|}{Accuracy(\%)}                                              &  66.67  &  80.00  &   73.33   &   86.67   &   80.00   &    86.67   &     \textbf{93.33}    \\ \bottomrule
\end{tabular}

\vspace{2mm}
\footnotesize \textbf{Note:} 
0 = \textit{Expired}, 1 = \textit{Moldy}, 2 = \textit{Normal}. 
Misclassified predictions are highlighted using the format \fbox{$x$}, where $x$ denotes the predicted class.
\label{tabular_online_testing}
\end{table*}



Firstly, after removing the FDEC, the overall accuracy dropped from 99.89\% to 85.44\%, a decrease of 14.45\%, representing the most pronounced performance degradation among all ablated variants. This indicated that FDEC played a central role in enhancing the dual-modal feature representations of olfactory and visual data. In particular, the Precision, Recall, and F1-Score of all three classes showed substantial declines, suggesting that without FDEC, the model was unable to effectively characterize the sample features.

Secondly, after removing the SE, the overall accuracy decreased by 5.67\%, indicating that the channel-wise recalibration provided by SE played an important role in highlighting effective olfactory features. A more noticeable performance drop occurred particularly in the expired and moldy classes, suggesting that SE helped capture the subtle response differences of gas-sensing materials when affected by mold. 

Thirdly, after removing the CBAM, the overall accuracy decreased by 5.00\%. Since mold spots typically exhibited sparse, small-scale, and spatially irregular distributions, the spatial attention offered by CBAM enabled the network to better focus on these key regions; therefore, its absence reduced the sensitivity of the visual branch to mold-related patterns.

Fourthly, after removing the SE and CBAM simultaneously, the model accuracy further dropped to 90.67\%, a decrease notably larger than that caused by removing either module alone. This indicated that the two submodules played complementary roles within FDRA: SE enhanced the discriminative capability of olfactory channel information, whereas CBAM strengthened the salience of visual spatial features. Their joint operation maximized the effectiveness of both visual and olfactory representations.

Removing any component from the proposed method resulted in a clear performance degradation as shown in Table \ref{table_ablation}, indicating that both FDEC and FDRA made substantive contributions to the final performance of the model. Among them, FDEC contributed the most significantly, as its removal led to the largest drop in accuracy, whereas the SE and CBAM submodules within FDRA provided complementary enhancements for the representations of the two modalities. The model achieved optimal performance only when all components worked collaboratively.

\begin{figure}[t!]
\centering
\includegraphics[width=3.4in]{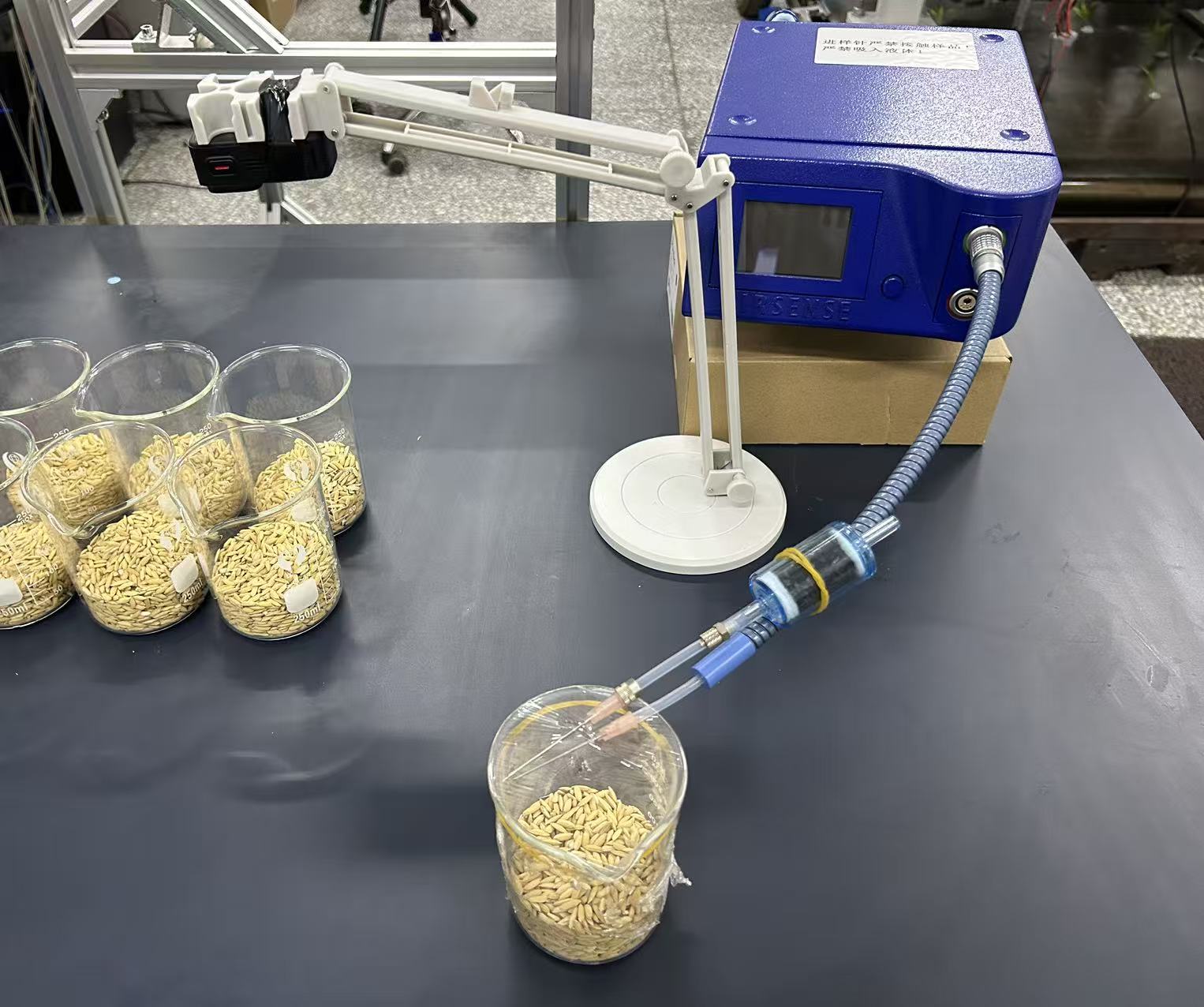}
\caption{Online detection setup for collecting RGB images and odor signals from rice samples.}
\label{online_testing}
\end{figure}

\subsection{Online Detection}
Experiments were conducted to validate the applicability of the proposed model in real rice deterioration detection scenarios, as shown in Fig. \ref{online_testing}. Fifteen rice samples were collected from an actual grain storage environment, and their sample types were consistent with those used in the offline experiments. The ground-truth labels of the samples were provided by an accredited inspection agency. For a more intuitive comparison, the prediction results of FDRA-Net and the aforementioned baseline models were presented in Table \ref{tabular_online_testing}.

It was observed that, under real granary conditions, the recognition performance of all models decreased compared with the offline experiments. Nevertheless, FDRA-Net still achieved the highest accuracy of 93.33\%, clearly outperforming all baseline models. From the perspective of class-wise recognition, the expired category was the most challenging under online conditions, as its visual degradation was often mild and the odor characteristics were not strongly distinguishable. As a result, multiple baseline models misclassified samples such as No. 5 and No. 14. In contrast, FDRA-Net only exhibited deviations on a few samples, indicating its stronger discriminative capability in weak-feature scenarios. Combined with the visualization analysis presented earlier, it was inferred that the FDRA module enhanced the class-discriminative weighting expression under online conditions, enabling the model to maintain stable predictions even when facing subtle differences such as weak mold or mild degradation.

Moreover, this study adopted a standard RGB camera for image acquisition rather than relying on costly and complex hyperspectral imaging devices, which rendered the proposed approach more practical and economical for real-world deployment. RGB cameras required neither professional imaging conditions nor complicated spectral calibration procedures, and they offered smaller device size and lower maintenance cost. Consequently, the proposed method significantly simplified the procedures while ensuring high recognition accuracy, providing a feasible solution for enhanced rice deterioration detection.



\section{Conclusion}
In this study, we propose a feature recalibration based olfactory–visual multimodal model for enhanced rice deterioration detection in real granary scenarios. The multimodal enhanced rice deterioration detection framework is established and validated with a focus on improving the reliability of data acquisition, the adequacy of feature representation, the robustness of model discrimination, and overall engineering practicality. FDEC reconstructs a labeled multimodal embedded-feature dataset and enhances sample representation. FDRA-Net recalibrates olfactory channel responses and emphasizes discriminative visual cues on the rice surface, improving sensitivity to fine-grained deterioration. As a result, the proposed method achieves 99.89\% accuracy in the offline experiments and maintains 93.33\% accuracy in the online detection. These results demonstrate that the proposed method effectively captures fine-grained degradation features and exhibits strong potential for practical deployment.

In further research, this work can be advanced from the perspectives of deployment feasibility, long-term robustness, and inference efficiency. First, for edge-side and real-time online applications, techniques such as model pruning, quantization, knowledge distillation, and operator-level optimization can be incorporated to further reduce model size and inference latency. Meanwhile, an integrated framework for data acquisition, inference, and alert generation can be refined to accommodate low-power, resource-constrained embedded platforms. Second, to address long-term operation and distribution drift in granary environments, sensor drift compensation and domain adaptation strategies can be introduced, including unlabeled data–based distribution alignment, online calibration, and few-shot incremental updates. These strategies aim to enhance generalization stability across seasons, storage facilities, and sample batches, and to further evaluate reliability under weak degradation conditions such as temperature and humidity fluctuations, background odor interference, and mild mold contamination. Third, from the perspectives of data enrichment and interpretability, the scale of online samples covering multiple rice varieties, storage periods, and operating conditions can be expanded, along with improvements to annotation and quality traceability. These enhanced interpretability analyses, in turn, improve the auditability and usability of the method in practical quality inspection workflows and facilitate its transfer and extension to the detection of other agrifood.

\section{Acknowledgments}
This work was supported by the National Key Research and Development Program Project (Grant No. 2023YFD2301604), the National Natural Science Foundation of China (Grant No. 62350710797 and No. 62103161), the Key Research and Development Program of Heilongjiang Province (Grant No. 2023ZX01A24 and No. JD2023GJ01-01), and the Project of Laboratory of Advanced Agricultural Sciences, Heilongjiang Province (Grant No. ZY04JD05-010). 
 
\bibliographystyle{IEEEtran}
\bibliography{refs}

\begin{IEEEbiography}[{\includegraphics[width=1in,height=1.25in,clip,keepaspectratio]{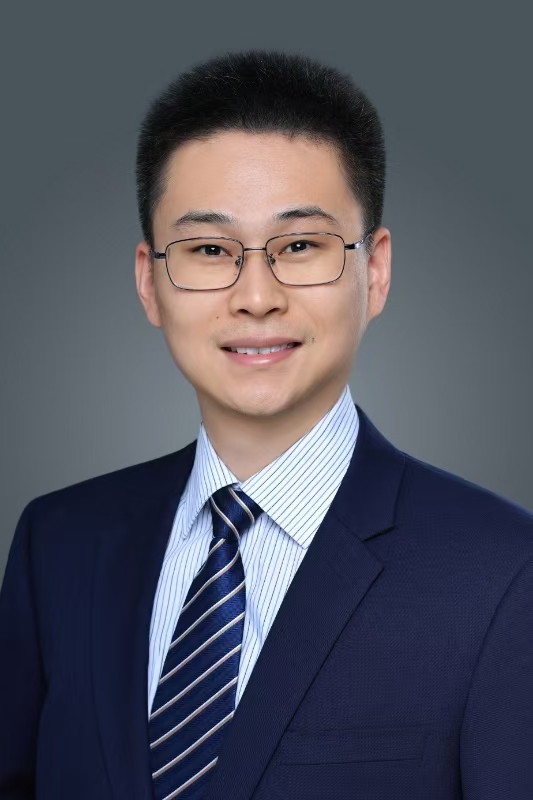}}]{Rongqiang Zhao}
received the Ph.D. degree in Harbin Institute of Technology, Harbin, China, in 2018. 

He is currently an Associate Professor with the Faculty of Computing, Harbin Institute of Technology, Harbin 150001, China, and also with the National Key Laboratory of Smart Farm Technologies and Systems, Harbin Institute of Technology, Harbin 150001, China.
\end{IEEEbiography}

\begin{IEEEbiography}[{\includegraphics[width=1in,height=1.25in,clip,keepaspectratio]{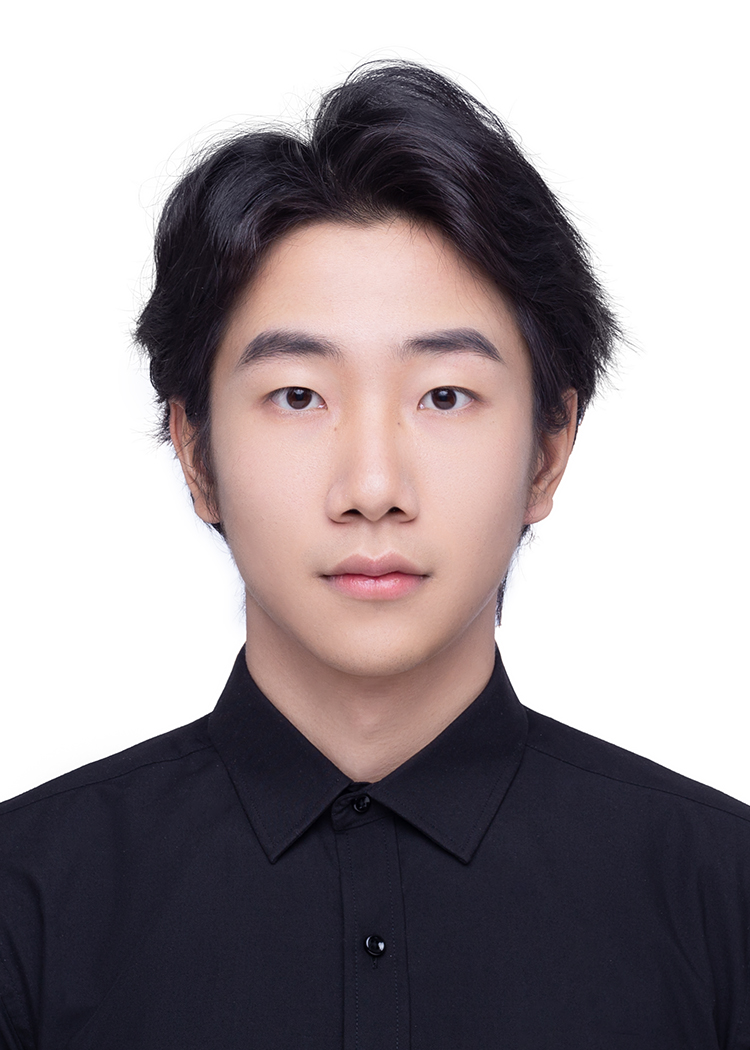}}]
{Hengrui Hu}
received the B.Eng. degree in Automation from the School of Future Technology, Harbin Institute of Technology, Harbin 150001, China, in 2025. 

He is currently pursuing the M.Sc. degree with the Faculty of Computing, Harbin Institute of Technology. His research interests include multimodal deep learning.
\end{IEEEbiography}

\begin{IEEEbiography}[{\includegraphics[width=1in,height=1.25in,clip,keepaspectratio]{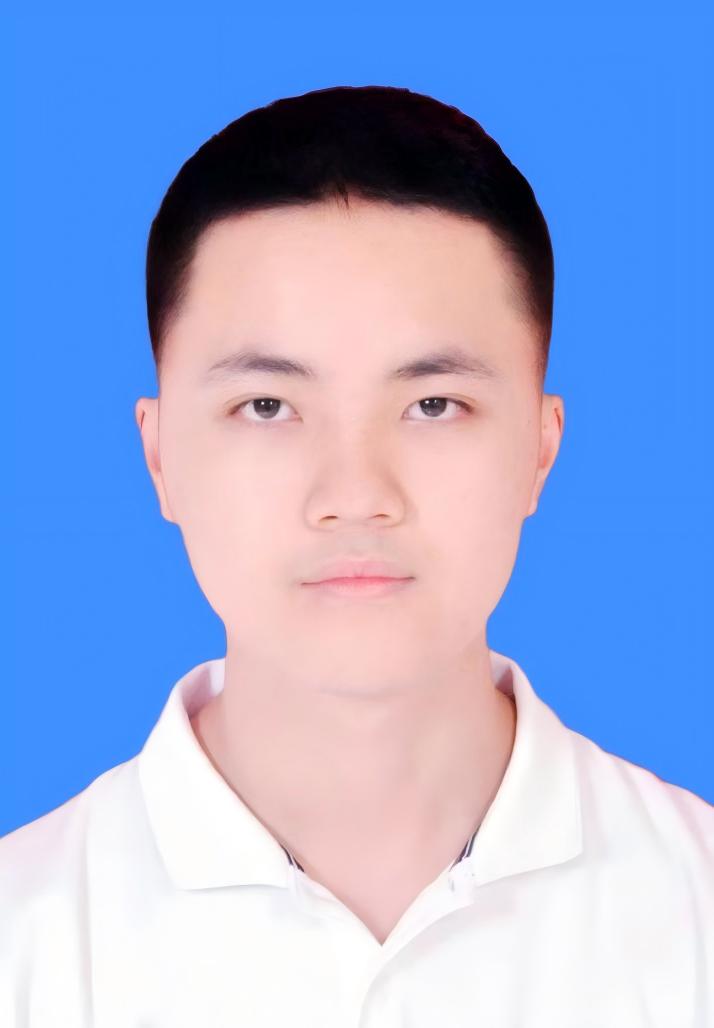}}]
{Yijing Wang}
received the B.Eng. degree in Computer Science and Technology from the School of Future Technology, Harbin Institute of Technology, Harbin 150001, China, in 2025.

He is currently pursuing the M.Sc. degree with the Faculty of Computing, Harbin Institute of Technology. His research interests include autonomous driving.
\end{IEEEbiography}

\begin{IEEEbiography}[{\includegraphics[width=1in,height=1.25in,clip,keepaspectratio]{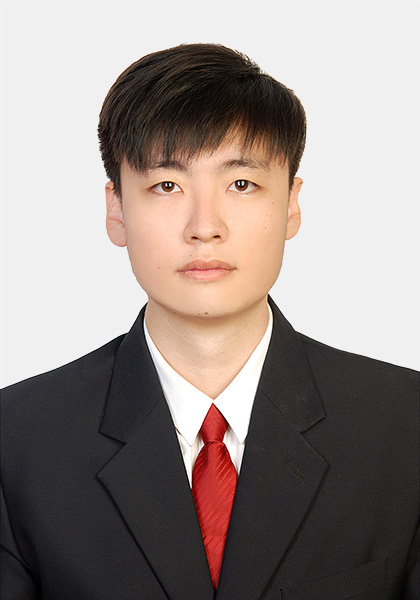}}]{Mingchun Sun}
received the B.S. and M.S. degrees in the School of Automation Engineering from the University of Electronic Science and Technology of China, Chengdu, China, in 2018, and 2021, respectively. 

He is currently pursuing the Ph.D. degree with the Harbin Institute of Technology, Harbin, China. His current research interests include data augmentation and deep learning on agricultural sensor datasets with limited samples (e-mail: sunmingchun@stu.hit.edu.cn).
\end{IEEEbiography}

\begin{IEEEbiography}[{\includegraphics[width=1in,height=1.25in,clip,keepaspectratio]{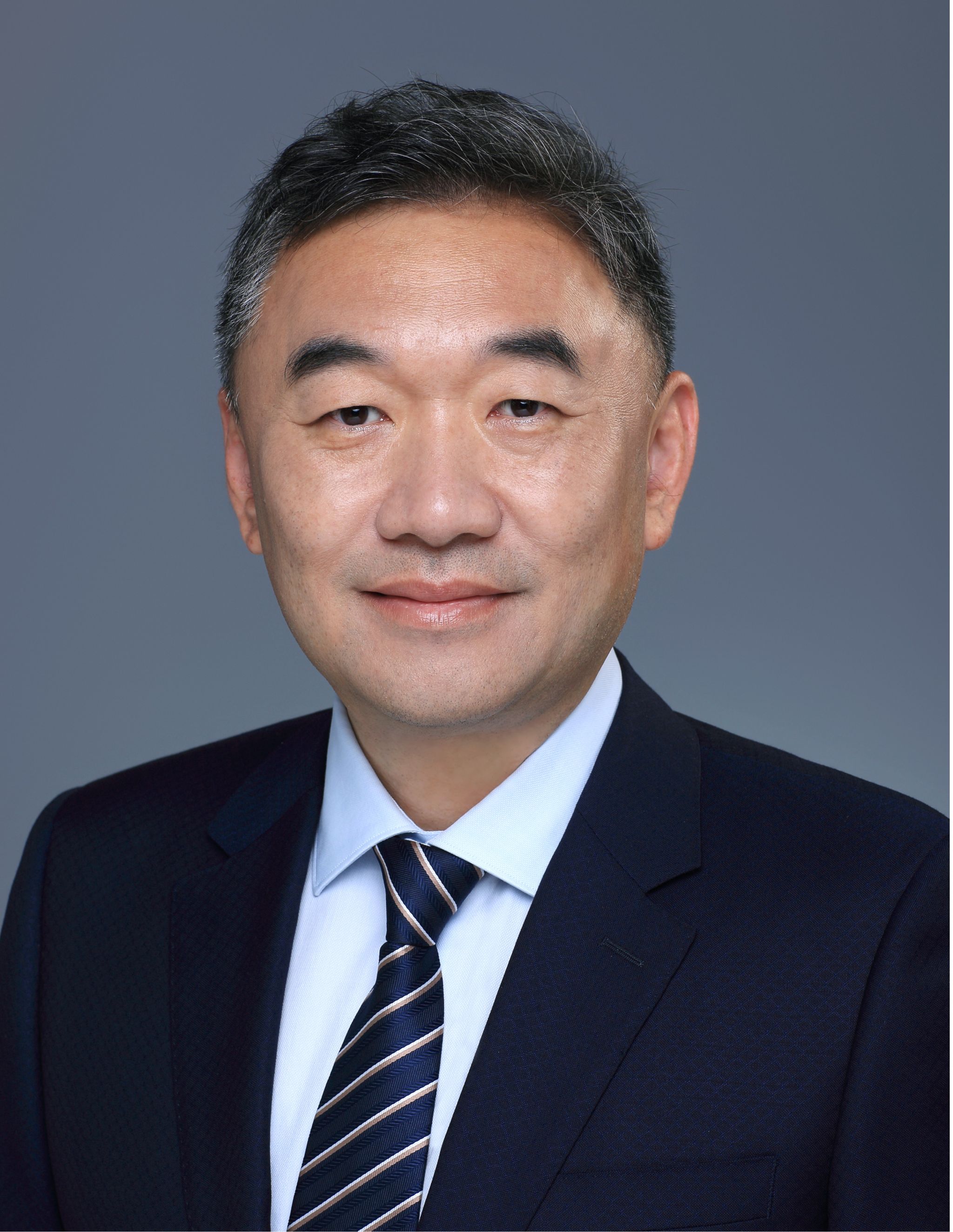}}]
{Jie Liu}
(Fellow, IEEE) received the Ph.D. degree in electrical engineering and computer science from the University of California at Berkeley, Berkeley, CA, USA, in 2001.

He is currently a Chair Professor with Harbin Institute of Technology, Shenzhen, China. His research interests include artificial intelligence, control engineering, the Internet of Things, and computer system.
\end{IEEEbiography}

\end{document}